\documentclass[]{interact}

\usepackage{epstopdf}
\usepackage[caption=false]{subfig}

\usepackage[numbers,sort&compress]{natbib}
\bibpunct[, ]{[}{]}{,}{n}{,}{,}
\renewcommand\bibfont{\fontsize{10}{12}\selectfont}
\makeatletter
\def\NAT@def@citea{\def\@citea{\NAT@separator}}
\makeatother

\usepackage{multirow}
\usepackage{color}
\usepackage{mathtools}
\usepackage{url}
\usepackage{graphicx}
\usepackage{algorithm}
\usepackage{algorithmic}

\newtheorem{theorem}{Theorem}[section]
\newtheorem{lemma}[theorem]{Lemma}

\begin{document}


\title{Iteration and Stochastic First-order Oracle Complexities of Stochastic Gradient Descent using Constant and Decaying Learning Rates}

\author{
Kento Imaizumi\textsuperscript{*1} and Hideaki Iiduka\textsuperscript{*1}\thanks{CONTACT Kento Imaizumi. Email: ee207005@meiji.ac.jp, Hideaki Iiduka. Emal: iiduka@cs.meiji.ac.jp}
\affil{\textsuperscript{*}Equal contribution; \textsuperscript{1}Department of Computer Science, Meiji University, Japan}
}

\maketitle

\begin{abstract}
The performance of stochastic gradient descent (SGD), which is the simplest first-order optimizer for training deep neural networks, depends on not only the learning rate but also the batch size. They both affect the number of iterations and the stochastic first-order oracle (SFO) complexity needed for training. In particular, the previous numerical results indicated that, for SGD using a constant learning rate, the number of iterations needed for training decreases when the batch size increases, and the SFO complexity needed for training is minimized at a critical batch size and that it increases once the batch size exceeds that size. Here, we study the relationship between batch size and the iteration and SFO complexities needed for nonconvex optimization in deep learning with SGD using constant or decaying learning rates and show that SGD using the critical batch size minimizes the SFO complexity. We also provide numerical comparisons of SGD with the existing first-order optimizers and show the usefulness of SGD using a critical batch size. Moreover, we show that measured critical batch sizes are close to the sizes estimated from our theoretical results.
\end{abstract}

\section{Introduction}
\subsection{Background}
\label{sec:1.1}
First-order optimizers can train deep neural networks by minimizing loss functions called the expected and empirical risks. They use stochastic first-order derivatives (stochastic gradients), which are estimated from the full gradient of the loss function. The simplest first-order optimizer is stochastic gradient descent (SGD) \citep{robb1951,zinkevich2003,nem2009,gha2012,gha2013}, which has a number of variants, including momentum variants\citep{polyak1964,nest1983} and numerous adaptive variants, such as adaptive gradient (AdaGrad) \citep{adagrad}, root mean square propagation (RMSProp) \citep{rmsprop}, adaptive moment estimation (Adam) \citep{adam}, adaptive mean square gradient (AMSGrad) \citep{reddi2018}, and Adam with decoupled weight decay (AdamW) \citep{loshchilov2018decoupled}. 

SGD can be applied to nonconvex optimization \citep{doi:10.1137/120880811,Ghadimi:2016aa,NEURIPS2019_2557911c,feh2020,chen2020,sca2020,loizou2021,wang2021on,Arjevani:2023aa,khaled2022better}, where its performance strongly depends on the learning rate $\alpha_k$. For example, under the bounded variance assumption, SGD using a constant learning rate $\alpha_k = \alpha$ satisfies $\frac{1}{K} \sum_{k=0}^{K-1} \|\nabla f(\bm{\theta}_k)\|^2 = O(\frac{1}{K}) + \sigma^2$ \citep[Theorem 12]{sca2020} and SGD using a decaying learning rate (i.e., $\alpha_k \to 0$) satisfies that $\frac{1}{K} \sum_{k=0}^{K-1} \mathbb{E}[\|\nabla f(\bm{\theta}_k)\|^2] = O(\frac{1}{\sqrt{K}})$ \citep[Theorem 11]{sca2020}, where $(\bm{\theta}_k)_{k\in\mathbb{N}}$ is the sequence generated by SGD to find a local minimizer of $f$, $K$ is the number of iterations, and $\sigma^2$ is the upper bound of the variance. 

The performance of SGD also depends on the batch size $b$. The convergence analyses reported in \citep{Ghadimi:2016aa,JMLR:v18:16-595,NIPS2011_b55ec28c,chen2020,Arjevani:2023aa} indicated that SGD with a decaying learning rate and large batch size converges to a local minimizer of the loss function. In \citep{l.2018dont}, it was numerically shown that using an enormous batch reduces both the number of parameter updates and model training time. Moreover, setting appropriate batch sizes for optimizers used in training generative adversarial networks were investigated in \citep{pmlr-v202-sato23b}.

\subsection{Motivation}
\label{sec:1.2}
The previous numerical results in \citep{shallue2019} indicated that, for SGD using constant or linearly decaying learning rates, the number of iterations $K$ needed to train a deep neural network decreases as the batch size $b$ increases. Motivated by the numerical results in \citep{shallue2019}, we decided to clarify {\em the theoretical iteration complexity} of SGD with a constant or decaying learning rate in training a deep neural network. We used the performance measure of previous theoretical analyses of SGD, i.e., $\min_{k\in [0:K-1]} \mathbb{E}[\|\nabla f(\bm{\theta}_k)\|] \leq \epsilon$, where $\epsilon$ $(>0)$ is the precision and $[0:K-1] := \{0,1,\ldots,K-1\}$. We found that, if SGD is an $\epsilon$--approximation, i.e., $\min_{k\in [0:K-1]} \mathbb{E}[\|\nabla f(\bm{\theta}_k)\|] \leq \epsilon$, then it can train a deep neural network in $K$ iterations.

In addition, the numerical results in \citep{shallue2019} indicated an interesting fact wherein diminishing returns exist beyond a critical batch size; i.e., the number of iterations needed to train a deep neural network does not strictly decrease beyond the critical batch size. Here, we define {\em the stochastic first-order oracle (SFO) complexity} as $N := Kb$, where $K$ is the number of iterations needed to train a deep neural network and $b$ is the batch size, as stated above. The deep neural network model uses $b$ gradients of the loss functions per iteration. The model has a stochastic gradient computation cost of $N = Kb$. From the numerical results in \citep[Figures 4 and 5]{shallue2019}, we can conclude that the critical batch size $b^\star$ (if it exists) is useful for SGD, since the SFO complexity $N(b)$ is minimized at $b = b^\star$ and the SFO complexity increases once the batch size exceeds $b^\star$. Hence, on the basis of the first motivation stated above, we decided to clarify the SFO complexities needed for SGD using a constant or decaying learning rate to be an $\epsilon$--approximation.

\subsection{Contribution}
\subsubsection{Upper bound of theoretical performance measure}
\label{sec:1.3.1}
To clarify the iteration and SFO complexities needed for SGD to be an $\epsilon$--approximation, we first give upper bounds of $\min_{k\in [0:K-1]} \mathbb{E}[\|\nabla f(\bm{\theta}_k)\|^2]$ for SGD to generate a sequence $(\bm{\theta}_k)_{k\in \mathbb{N}}$ with constant or decaying learning rates (see Theorem \ref{theorem:1} for the definitions of $C_i$ and $D_i$). As our aim is to show that SGD is an $\epsilon$--approximation $\min_{k\in [0:K-1]} \mathbb{E}[\|\nabla f(\bm{\theta}_k)\|^2] \leq \epsilon^2$, it is desirable that the upper bounds of $\min_{k\in [0:K-1]} \mathbb{E}[\|\nabla f(\bm{\theta}_k)\|^2]$ be small. Table \ref{table:0} indicates that the upper bounds become small when the number of iterations and batch size are large. The table also indicates that the convergence of SGD strongly depends on the batch size, since the variance terms (including $\sigma^2$ and $b$; see Theorem \ref{theorem:1} for the definitions of $C_2$ and $D_2$) in the upper bounds of $\min_{k\in [0:K-1]} \mathbb{E}[\|\nabla f(\bm{\theta}_k)\|^2]$ decrease as the batch size becomes larger.

\begin{table*}[ht]
\centering
\caption{Upper bounds of $\min_{k\in [0:K-1]} \mathbb{E}[\|\nabla f(\bm{\theta}_k)\|^2]$ for SGD using a constant or decaying learning rate and the critical batch size to minimize the SFO complexities and achieve $\min_{k\in [0:K-1]} \mathbb{E}[\|\nabla f (\bm{\theta}_k)\|] \leq \epsilon$ ($C_i$ and $D_i$ are positive constants, $K$ is the number of iterations, $b$ is the batch size, $T \geq 1$, $\epsilon > 0$, and $L$ is the Lipschitz constant of $\nabla f$)}\label{table:0}
\begin{tabular}{llll}
\toprule
\multicolumn{2}{l}{Learning Rate} & Upper Bound & Critical Batch Size\\
\midrule
\multicolumn{2}{l}{Constant $\alpha \in (0,\frac{2}{L})$} 
& $\displaystyle{\frac{C_1}{K} + \frac{C_2}{b}}$ 
& $\displaystyle{\frac{2 C_2}{\epsilon^2}}$ \\
\midrule
\multirow{3}{*}{}
& $a \in (0,\frac{1}{2})$ 
& $\displaystyle{\frac{D_1}{K^a} + \frac{D_2}{(1-2a)K^a b}}$ 
& $\displaystyle{\frac{(1-a)D_2}{a(1-2a)D_1}}$ \\
\cmidrule{2-4}
Decay & $a = \frac{1}{2}$ 
& $\displaystyle{\frac{D_1}{\sqrt{K}} + \left(\frac{1}{\sqrt{K}} + 1 \right) \frac{D_2}{b}}$ 
& $\displaystyle{\approx \frac{D_2}{\epsilon^2}}$ \\
\cmidrule{2-4}
$\alpha_k = \frac{1}{\left(\left\lfloor\frac{k}{T}\right\rfloor+1\right)^a}$ 
& $a \in (\frac{1}{2},1)$ 
& $\displaystyle{\frac{D_1}{K^{1-a}} + \frac{2 a D_2}{(2a -1)K^{1-a} b}}$ 
& $\displaystyle{\frac{2 a^2 D_2}{(1-a)(2a-1)D_1}}$ \\
\bottomrule 
\end{tabular}
\end{table*}

\subsubsection{Critical batch size to reduce SFO complexity}
\label{sec:1.3.2}
Section \ref{sec:1.3.1} showed that using large batches is appropriate for SGD in the sense of minimizing the upper bound of the performance measure. Here, we are interested in finding appropriate batch sizes from the viewpoint of the computation cost. This is because the SFO complexity increases with the batch size. As indicated in Section \ref{sec:1.2}, the critical batch size $b^\star$ minimizes the SFO complexity, $N = Kb$. Hence, we will investigate the properties of the SFO complexity $N = Kb$ needed to achieve an $\epsilon$--approximation. Here, let us consider SGD using a constant learning rate. From the ``Upper Bound" row in Table \ref{table:0}, we have 
\begin{align*}
\min_{k\in [0:K-1]} \mathbb{E}[\|\nabla f(\bm{\theta}_k)\|^2] 
\leq \underbrace{\frac{C_1}{K} + \frac{C_2}{b} \leq \epsilon^2}_{\Leftrightarrow K \geq K(b) := \frac{C_1 b}{\epsilon^2 b - C_2} \text{ } \left(b > \frac{C_2}{\epsilon^2} \right)}.
\end{align*}
We can check that the number of iterations, $K(b) := \frac{C_1 b}{\epsilon^2 b - C_2}$, needed to achieve an $\epsilon$--approximation is monotone decreasing and convex with respect to the batch size  (Theorem \ref{theorem:2}). Accordingly, we have that $K(b) \geq \inf \{ K \colon \min_{k\in [0:K-1]} \mathbb{E}[\|\nabla f(\bm{\theta}_k)\|] \leq \epsilon \}$, where SGD using the batch size $b$ generates $(\bm{\theta}_k)_{k=0}^{K-1}$. Moreover, we find that the SFO complexity is $N(b) = K(b) b = \frac{C_1 b^2}{\epsilon^2 b - C_2}$. The convexity of $N(b) = \frac{C_1 b^2}{\epsilon^2 b - C_2}$ (Theorem \ref{theorem:3}) ensures that a critical batch size $b^\star = \frac{2 C_2}{\epsilon^2}$ whereby $N'(b^\star) = 0$ exists such that $N(b)$ is minimized at $b^\star$ (see the ``Critical Batch Size" row in Table \ref{table:0}). A similar discussion guarantees the existence of a critical batch size for SGD using a decaying learning rate $\alpha_k = \frac{1}{\left(\left\lfloor\frac{k}{T}\right\rfloor+1\right)^a}$, where $T \geq 1$, $a \in (0,\frac{1}{2}) \cup (\frac{1}{2},1)$, and $\lfloor\cdot\rfloor$ is the floor function (see the ``Critical Batch Size" row in Table \ref{table:0}).

Meanwhile, for a decaying learning rate $\alpha_k = \frac{1}{\sqrt{\left\lfloor\frac{k}{T}\right\rfloor+1}}$, although $N(b)$ is convex with respect to $b$, we have that $N'(b) > 0$ for all $b > \frac{D_2}{\epsilon^2}$ (Theorem \ref{theorem:3}(iii)). Hence, for this case, a critical batch size $b^\star$ defined by $N'(b^\star) = 0$ does not exist. However, since the critical batch size minimizes the SFO complexity $N$, we can define one as follows: $b^\star \approx \frac{D_2}{\epsilon^2}$. Accordingly, we have that $N(b^\star) \geq \inf \{ Kb \colon \min_{k\in [0:K-1]} \mathbb{E}[\|\nabla f(\bm{\theta}_k)\|] \leq \epsilon \}$, where SGD using $b^\star$ generates $(\bm{\theta}_k)_{k=0}^{K-1}$.

\subsubsection{Iteration and SFO complexities}
\label{sec:1.3.3}
Let $\mathcal{F}(n, \Delta_0, L)$ be an $L$--smooth function class with $f := \frac{1}{n} \sum_{i=1}^n f_i$ and $f(\bm{\theta}_0) - f_\star \leq \Delta_0$ (see (C1)) and let $\mathcal{O}(b,\sigma^2)$ be a stochastic first-order oracle class (see (C2) and (C3)). The iteration complexity $\mathcal{K}_\epsilon$ \citep[(7)]{Arjevani:2023aa} and SFO complexity $\mathcal{N}_\epsilon$ needed for SGD to be an $\epsilon$--approximation are defined as 
\begin{align}\label{K_N}
&\mathcal{K}_\epsilon (n, b, \alpha_k, \Delta_0, L, \sigma^2) := 
\sup_{\mathsf{O} \in \mathcal{O}(b,\sigma^2)} \sup_{f \in \mathcal{F}(n, \Delta_0, L)}
\inf
\left\{ K \colon \min_{k\in [0:K-1]}\mathbb{E}[\|\nabla f(\bm{\theta}_k)\|] \leq \epsilon \right\},\\
&\mathcal{N}_\epsilon (n, b, \alpha_k, \Delta_0, L, \sigma^2) 
:= \mathcal{K}_\epsilon (n, b, \alpha_k, \Delta_0, L, \sigma^2) b.
\end{align}
Table \ref{table:1} summarizes the iteration and SFO complexities (see also Theorem \ref{theorem:4}). Corollaries 6 and 7 in \citep{Ghadimi:2016aa} are the same as our results for SGD with a constant learning rate in Theorems \ref{theorem:1} and \ref{theorem:3}, since the randomized stochastic projected gradient free algorithm in \citep{Ghadimi:2016aa} which is a stochastic zeroth-order (SZO) method that coincides with SGD and it can be applied to the situations where only noisy function values are available. In particular, Corollary 6 in \citep{Ghadimi:2016aa} gave the convergence rate of the SZO methods using a fixed batch size, and Corollary 7 indicated the SZO complexity of the SZO method is the same as the SFO complexity. Hence, Corollaries 6 and 7 in \citep{Ghadimi:2016aa} lead to the finding that the iteration complexity of SGD using a constant learning rate is $O(1/\epsilon^2)$ and the SFO complexity of SGD using a constant learning rate is $O(1/\epsilon^4)$.

Since the positive constants $C_i$ and $D_i$ depend on the learning rate, we need to compare numerically the performance of SGD with a constant learning rate with that of SGD with a decaying learning rate. Moreover, we also need to compare SGD with the existing first-order optimizers in order to verify its usefulness. Section \ref{sec:4} presents numerical comparisons showing that SGD using the critical batch size outperforms the existing first-order optimizers. We also show that the measured critical batch sizes are close to the theoretical sizes.

\begin{table*}[ht]
\centering
\caption{Iteration and SFO complexities needed for SGD using a constant or decaying learning rate to be an $\epsilon$--approximation (The critical batch sizes are used to compute $\mathcal{K}_\epsilon$ and $\mathcal{N}_\epsilon$)}\label{table:1}
\begin{tabular}{llll}
\toprule
\multicolumn{2}{l}{Learning Rate} & Iteration Complexity $\mathcal{K}_\epsilon$ & SFO Complexity $\mathcal{N}_\epsilon(n, b, \alpha_k, \Delta_0, L, \sigma^2)$\\
\midrule
\multicolumn{2}{l}{Constant $\alpha \in (0,\frac{2}{L})$} 
& $\displaystyle{O \left(\frac{1}{\epsilon^2} \right) = \sup_{f,\mathsf{O}} K(b^\star)}$ 
& $\displaystyle{O \left(\frac{1}{\epsilon^4} \right) = \sup_{f,\mathsf{O}} \frac{4 C_1 C_2}{\epsilon^4}}$ \\
\midrule
\multirow{3}{*}{} 
& $a \in (0,\frac{1}{2})$ 
& $\displaystyle{O \left(\frac{1}{\epsilon^{\frac{2}{a}}} \right) = \sup_{f,\mathsf{O}} K(b^\star)}$ 
& $\displaystyle{O \left(\frac{1}{\epsilon^{\frac{2}{a}}} \right) = \sup_{f,\mathsf{O}} \frac{(1- a)^{1 - \frac{1}{a}} D_2}{a (1 - 2a)D_1^{1 - \frac{1}{a}} \epsilon^{\frac{2}{a}}}}$ \\
\cmidrule{2-4}
Decay & $a = \frac{1}{2}$ 
& $\displaystyle{O \left(\frac{1}{\epsilon^4} \right) = \sup_{f,\mathsf{O}} K(b^\star)}$ 
& $\displaystyle{O \left(\frac{1}{\epsilon^{6}} \right) = \sup_{f,\mathsf{O}} \left( \frac{D_1(D_2+1)}{\epsilon^2} + D_2\right)^2\frac{D_1+1}{\epsilon^2}}$ \\
\cmidrule{2-4}
$\alpha_k = \frac{1}{\left(\left\lfloor\frac{k}{T}\right\rfloor+1\right)^a}$ 
& $a \in (\frac{1}{2},1)$ 
& $\displaystyle{O \left(\frac{1}{\epsilon^{\frac{2}{1-a}}} \right) = \sup_{f,\mathsf{O}} K(b^\star)}$ 
& $\displaystyle{O \left(\frac{1}{\epsilon^{\frac{2}{1-a}}} \right) = 
\sup_{f,\mathsf{O}} \frac{2a^{2 - \frac{1}{1-a}} (1-a)^{-1} D_2}{(2a-1)D_1^{1 - \frac{1}{1-a}} \epsilon^{\frac{2}{1-a}}}}$ \\
\bottomrule 
\end{tabular}
\end{table*}

\section{Nonconvex Optimization and SGD}
\label{sec:2}
\subsection{Nonconvex optimization in deep learning}
\label{subsec:2.1}
Let $\mathbb{R}^d$ be a $d$-dimensional Euclidean space with inner product $\langle \bm{x},\bm{y} \rangle := \bm{x}^\top \bm{y}$ inducing the norm $\| \bm{x}\|$ and $\mathbb{N}$ be the set of nonnegative integers. Define $[0:n] := \{0,1,\ldots,n\}$ for $n \geq 1$. Let $(x_k)_{k\in\mathbb{N}}$ and $(y_k)_{k\in\mathbb{N}}$ be positive real sequences and let $x(\epsilon), y(\epsilon) > 0$, where $\epsilon > 0$. $O$ denotes Landau's symbol; i.e., $y_k = O(x_k)$ if there exist $c > 0$ and $k_0 \in \mathbb{N}$ such that $y_k \leq c x_k$ for all $k \geq k_0$, and $y(\epsilon) = O (x(\epsilon))$ if there exists $c > 0$ such that $y(\epsilon) \leq cx(\epsilon)$. Given a parameter $\bm{\theta} \in \mathbb{R}^d$ and a data point $z$ in a data domain $Z$, a machine-learning model provides a prediction whose quality can be measured in terms of a differentiable nonconvex loss function $\ell(\bm{\theta};z)$. We aim to minimize the empirical loss defined for all $\bm{\theta} \in \mathbb{R}^d$ by $f(\bm{\theta}) = \frac{1}{n} \sum_{i=1}^n \ell(\bm{\theta};z_i) = \frac{1}{n} \sum_{i=1}^n f_i(\bm{\theta})$, where $S = (z_1, z_2, \ldots, z_n)$ denotes the training set (We assume that the number of training data $n$ is large) and $f_i (\cdot) := \ell(\cdot;z_i)$ denotes the loss function corresponding to the $i$-th training data $z_i$.

\subsection{SGD}
\label{subsec:2.2}
\subsubsection{Conditions and algorithm}
We assume that a stochastic first-order oracle (SFO) exists such that, for a given $\bm{\theta} \in \mathbb{R}^d$, it returns a stochastic gradient $\mathsf{G}_{\xi}(\bm{\theta})$ of the function $f$, where a random variable $\xi$ is independent of $\bm{\theta}$. Let $\mathbb{E}_\xi [\cdot]$ be the expectation taken with respect to $\xi$. The following are standard conditions. 
\begin{enumerate}
\item[(C1)] $f := \frac{1}{n} \sum_{i=1}^n f_i \colon \mathbb{R}^d \to \mathbb{R}$ is $L$--smooth, i.e., $\nabla f \colon \mathbb{R}^d \to \mathbb{R}^d$ is $L$--Lipschitz continuous (i.e., $\|\nabla f (\bm{x}) - \nabla f(\bm{y})\| \leq L \|\bm{x} - \bm{y} \|$). $f$ is bounded below from $f_\star \in \mathbb{R}$. Let $\Delta_0 > 0$ satisfy $f(\bm{\theta}_0) - f_\star \leq \Delta_0$, where $\bm{\theta}_0$ is an initial point.
\item[(C2)] Let $(\bm{\theta}_k)_{k\in \mathbb{N}} \subset \mathbb{R}^d$ be the sequence generated by SGD. For each iteration $k$, $\mathbb{E}_{\xi_k} [ \mathsf{G}_{\xi_k}(\bm{\theta}_k) ] = \nabla f(\bm{\theta}_k)$, where $\xi_0, \xi_1, \ldots$ are independent samples and the random variable $\xi_k$ is independent of $(\bm{\theta}_l)_{l=0}^k$. There exists a nonnegative constant $\sigma^2$ such that $\mathbb{E}_{\xi_k} [ \|\mathsf{G}_{\xi_k}(\bm{\theta}_k) - \nabla f(\bm{\theta}_k) \|^2 ] \leq \sigma^2$.
\item[(C3)] For each iteration $k$, SGD samples a batch $B_{k}$ of size $b$ independently of $k$ and estimates the full gradient $\nabla f$ as $\nabla f_{B_k} (\bm{\theta}_k) := \frac{1}{b} \sum_{i\in [b]} \mathsf{G}_{\xi_{k,i}}(\bm{\theta}_k)$, where $b \leq n$ and $\xi_{k,i}$ is a random variable generated by the $i$-th sampling in the $k$-th iteration.
\end{enumerate}

Algorithm \ref{algo:1} is the SGD optimizer under (C1)--(C3).

\begin{algorithm} 
\caption{SGD} 
\label{algo:1} 
\begin{algorithmic}[1] 
\REQUIRE
$\alpha_k \in (0,+\infty)$ (learning rate), $b \geq 1$ (batch size), $K \geq 1$ (iteration)
\ENSURE
$\bm{\theta}_{K}$
\STATE
$k \gets 0$, $\bm{\theta}_{0} \in\mathbb{R}^d$
\LOOP 
\STATE
$\nabla f_{B_k} (\bm{\theta}_k)
:= \frac{1}{b} \sum_{i\in [b]} \mathsf{G}_{\xi_{k,i}}(\bm{\theta}_k)$
\STATE 
$\bm{\theta}_{k+1} := \bm{\theta}_k - \alpha_k \nabla f_{B_k} (\bm{\theta}_k)$
\STATE $k \gets k+1$
\ENDLOOP 
\end{algorithmic}
\end{algorithm}

\subsubsection{Learning rates}
\label{subsec:2.2.3}
We use the following learning rates:
\begin{description}
\item[(Constant rate)] $\alpha_k$ does not depend on $k \in \mathbb{N}$, i.e., $\alpha_k = \alpha < \frac{2}{L}$ $(k \in \mathbb{N})$, where the upper bound $\frac{2}{L}$ of $\alpha$ is needed to analyze SGD (see Appendix \ref{a_1}). 
\item[(Decaying rate)] $(\alpha_k)_{k\in \mathbb{N}} \subset (0,+\infty)$ is monotone decreasing for $k$ (i.e., $\alpha_k \geq \alpha_{k+1}$) and converges to $0$. In particular, we will use $\alpha_k = \frac{1}{\left(\left\lfloor\frac{k}{T}\right\rfloor+1\right)^a}$, where $\textbf{(Decay 1) } a \in (0,\frac{1}{2}) \lor \textbf{ (Decay 2) } a = \frac{1}{2} \lor \textbf{ (Decay 3) } a \in (\frac{1}{2},1)$. It is guaranteed that there exists $k_0 \in \mathbb{N}$ such that, for all $k \geq k_0$, $\alpha_k < \frac{2}{L}$. Furthermore, we assume that $k_0 = 0$, since we can replace $\alpha_k$ with $\frac{\alpha}{\left(\left\lfloor\frac{k}{T}\right\rfloor+1\right)^a} \leq \alpha < \frac{2}{L}$ $(k\in\mathbb{N})$, where $\alpha \in (0,\frac{2}{L})$ is defined as in {\bf (Constant)}.
\end{description}

\section{Our Results}
\subsection{Upper bound of the squared norm of the full gradient}
Here, we give an upper bound of $\min_{k \in [0:K-1]} \mathbb{E} [\| \nabla f(\bm{\theta}_k) \|^2 ]$, where $\mathbb{E}[\cdot]$ stands for the total expectation, for the sequence generated by SGD using each of the learning rates defined in Section \ref{subsec:2.2.3}.

\begin{theorem}
[Upper bound of the squared norm of the full gradient]\label{theorem:1} The sequence $(\bm{\theta}_k)_{k\in \mathbb{N}}$ generated by Algorithm \ref{algo:1} under (C1)--(C3) satisfies that, for all $K \geq 1$,
\begin{align*}
&\min_{k \in [0:K-1]} \mathbb{E}\left[\| \nabla f(\bm{\theta}_k) \|^2 \right]
&\leq
\begin{dcases}
\displaystyle{\frac{C_1}{K} + \frac{C_2}{b}} &\textbf{ (Constant)}\\
\displaystyle{\frac{D_1}{K^a} + \frac{D_2}{(1-2a)K^a b}} &\textbf{ (Decay 1)}\\
\displaystyle{\frac{D_1}{\sqrt{K}} + \left(\frac{1}{\sqrt{K}} + 1 \right) \frac{D_2}{b}} &\textbf{ (Decay 2)}\\
\displaystyle{\frac{D_1}{K^{1-a}} + \frac{2 a D_2}{(2a -1)K^{1-a} b}} &\textbf{ (Decay 3)}
\end{dcases}
\end{align*}
where 
\begin{align*}
&C_1 := \frac{2(f(\bm{\theta}_0) - f_\star)}{(2 - L \alpha) \alpha},
\text{ } 
C_2 := \frac{L \sigma^2 \alpha}{2 - L \alpha}, \\
&D_1 := 
\displaystyle{\frac{2\left(f(\bm{\theta}_0)-f_\star\right)}{\alpha(2-L\alpha)}},
\text{ }
D_2 := \frac{T\alpha^2L\sigma^2}{2-L\alpha}.
\end{align*}
\end{theorem}

Theorem \ref{theorem:1} indicates that the upper bound of $\min_{k \in [0:K-1]} \mathbb{E}[\| \nabla f(\bm{\theta}_k) \|^2 ]$ consists of a bias term including $f(\bm{\theta}_0) - f_\star$ and a variance term including $\sigma^2$ and that these terms become small when the number of iterations and the batch size are large. In particular, the bias term using \textbf{(Constant)} is $O(\frac{1}{K})$, which is a better rate than using \textbf{(Decay 1)}--\textbf{(Decay 3)}.

\subsection{Number of iterations needed for SGD to be an $\epsilon$--approximation}
Let us suppose that SGD is an $\epsilon$--approximation defined as follows:
\begin{align}\label{e_app}
\mathbb{E}\left[\| \nabla f(\bm{\theta}_{K^*}) \|^2 \right] 
&:= \min_{k \in [0:K-1]} \mathbb{E}\left[\| \nabla f(\bm{\theta}_k) \|^2 \right] \leq \epsilon^2,
\end{align}
where $\epsilon > 0$ is the precision and $K^* \in [0:K-1]$. Condition (\ref{e_app}) implies that $\mathbb{E}[\| \nabla f(\bm{\theta}_{K^*}) \| ] \leq \epsilon$. Theorem \ref{theorem:1} below gives the number of iterations needed to be an $\epsilon$--approximation (\ref{e_app}).

\begin{theorem}
[Numbers of iterations needed for nonconvex optimization using SGD]\label{theorem:2} Let $(\bm{\theta}_k)_{k\in \mathbb{N}}$ be the sequence generated by Algorithm \ref{algo:1} under (C1)--(C3) and let $K \colon \mathbb{R} \to \mathbb{R}$ be 
\begin{align*}
K (b) = 
\begin{cases}
\displaystyle{\frac{C_1 b}{\epsilon^2 b - C_2}} &\textbf{ (Constant)}\\
\displaystyle{\left\{ \frac{1}{\epsilon^2} \left(\frac{D_2}{(1-2a)b} + D_1 \right) \right\}^{\frac{1}{a}}} &\textbf{ (Decay 1)}\\
\displaystyle{\left( \frac{D_1 b + D_2}{\epsilon^2 b - D_2} \right)^{2}} &\textbf{ (Decay 2)}\\
\displaystyle{\left\{ \frac{1}{\epsilon^2} \left(\frac{2 a D_2}{(2a -1)b} + D_1 \right) \right\}^{\frac{1}{1-a}}} &\textbf{ (Decay 3)}
\end{cases}
\end{align*}
where $C_1$, $C_2$, $D_1$ $(> \epsilon^2)$, and $D_2$ are defined as in Theorem \ref{theorem:1}, the domain of $K$ in {\em \textbf{(Constant)}} is $b > \frac{C_2}{\epsilon^2}$, and the domain of $K$ in {\em \textbf{(Decay 2)}} is $b > \frac{D_2}{\epsilon^2}$. Then, we have the following: 
\begin{enumerate}
\item[{\em (i)}] The above $K$ achieves an $\epsilon$--approximation (\ref{e_app}). 
\item[{\em (ii)}]The above $K$ is a monotone decreasing and convex function with respect to the batch size $b$.
\end{enumerate}
\end{theorem}

Theorem \ref{theorem:2} indicates that the number of iterations needed for SGD using constant or decay learning rates to be an $\epsilon$--approximation is small when the batch size is large. Hence, it is appropriate to set a large batch size in order to minimize the iterations needed for an $\epsilon$--approximation (\ref{e_app}). However, the SFO complexity, which is the cost of the stochastic gradient computation, grows larger with $b$. Hence, the appropriate batch size should also minimize the SFO complexity. 

\subsection{SFO complexity needed for SGD to be an $\epsilon$--approximation}
Theorem \ref{theorem:2} leads to the following theorem on the properties of the SFO complexity $N$ needed for SGD to be an $\epsilon$--approximation (\ref{e_app}).

\begin{theorem}
[SFO complexity needed for nonconvex optimization of SGD]\label{theorem:3} Let $(\bm{\theta}_k)_{k\in \mathbb{N}}$ be the sequence generated by Algorithm \ref{algo:1} under (C1)--(C3) and define $N \colon \mathbb{R} \to \mathbb{R}$ by 
\begin{align*}
N (b) 
&= K (b) b
=
\begin{cases}
\displaystyle{\frac{C_1 b^2}{\epsilon^2 b - C_2}} &\textbf{ (Constant)}\\
\displaystyle{\left\{ \frac{1}{\epsilon^2} \left(\frac{D_2}{(1-2a)b} + D_1 \right) \right\}^{\frac{1}{a}}} b &\textbf{ (Decay 1)}\\
\displaystyle{\left( \frac{D_1 b + D_2}{\epsilon^2 b - D_2} \right)^{2} b} &\textbf{ (Decay 2)}\\
\displaystyle{\left\{ \frac{1}{\epsilon^2} \left(\frac{2 a D_2}{(2a -1)b} + D_1 \right) \right\}^{\frac{1}{1-a}}} b &\textbf{ (Decay 3)}
\end{cases}
\end{align*}
where $C_1$, $C_2$, $D_1$ and $D_2$ are as in Theorem \ref{theorem:1}, the domain of $N$ in {\em \textbf{(Constant)}} is $b > \frac{C_2}{\epsilon^2}$, and the domain of $N$ in {\em \textbf{(Decay 2)}} is $b > \frac{D_2}{\epsilon^2}$. Then, we have the following:
\begin{enumerate}
\item[{\em (i)}] The above $N$ is convex with respect to the batch size $b$.
\item[{\em (ii)}] There exists a critical batch size
\begin{align}\label{cbs}
b^\star = 
\begin{cases}
\displaystyle{\frac{2 C_2}{\epsilon^2}} &\textbf{ (Constant)}\\
\displaystyle{\frac{(1-a) D_2}{a(1 - 2a)D_1}} &\textbf{ (Decay 1)}\\
\displaystyle{\frac{2 a^2 D_2}{(1-a)(2a -1)D_1}} &\textbf{ (Decay 3)}
\end{cases}
\end{align}
satisfying $N'(b^\star) = 0$ such that $b^\star$ minimizes the SFO complexity $N$. 
\item[{\em (iii)}] For {\em \textbf{(Decay 2)}}, $N'(b) > 0$ holds for all $b > \dfrac{D_2}{\epsilon^2}$.
\end{enumerate}
\end{theorem}

Theorem \ref{theorem:3}(ii) indicates that, if we can set a critical batch size (\ref{cbs}) for each of \textbf{(Constant)}, \textbf{(Decay 1)}, and \textbf{(Decay 3)}, then the SFO complexity will be minimized. However, it would be difficult to set $b^\star$ in (\ref{cbs}) before implementing SGD, since $b^\star$ in (\ref{cbs}) involves unknown parameters, such as $L$ and $\sigma^2$ (computing $L$ is NP-hard \citep{NEURIPS2018_d54e99a6}). Hence, we would like to estimate the critical batch sizes by using Theorem \ref{theorem:3}(ii) and (iii) (see Section \ref{sec:4.3}). Theorem \ref{theorem:3}(ii) indicates that the smaller $\epsilon$ is, the larger the critical batch size $b^{\star}$ in \textbf{(Constant)} becomes. Theorem \ref{theorem:3}(iii) indicates that the critical batch size is close to $\frac{D_2}{\epsilon^2}$ when using \textbf{(Decay 2)} to minimize the SFO complexity $N$.

\subsection{Iteration and SFO complexities of SGD}
Theorems \ref{theorem:2} and \ref{theorem:3} lead to the following theorem indicating the iteration and SFO complexities needed for SGD to be an $\epsilon$--approximation (see also Table \ref{table:1}). 

\begin{theorem}
[Iteration and SFO complexities of SGD]\label{theorem:4} The iteration and SFO complexities such that Algorithm \ref{algo:1} under (C1)--(C3) is an $\epsilon$--approximation (\ref{e_app}) are as follows:
\begin{align*}
&(\mathcal{K}_\epsilon (n, b^\star, \alpha_k, \Delta_0, L, \sigma^2),
\mathcal{N}_\epsilon (n, b^\star, \alpha_k, \Delta_0, L, \sigma^2))
=
\begin{cases}
\displaystyle{\left( O \left( \frac{1}{\epsilon^2} \right), O \left( \frac{1}{\epsilon^4} \right) \right)} &\textbf{ (Constant)}\\
\displaystyle{\left( O \left( \frac{1}{\epsilon^{\frac{2}{a}}} \right), O \left( \frac{1}{\epsilon^{\frac{2}{a}}} \right) \right)} &\textbf{ (Decay 1)}\\
\displaystyle{\left( O \left( \frac{1}{\epsilon^4} \right), O \left( \frac{1}{\epsilon^{6}} \right) \right)} &\textbf{ (Decay 2)}\\
\displaystyle{\left( O \left( \frac{1}{\epsilon^{\frac{2}{1-a}}} \right), O \left( \frac{1}{\epsilon^{\frac{2}{1-a}}} \right) \right)} &\textbf{ (Decay 3)}
\end{cases}
\end{align*}
where $\mathcal{K}_\epsilon (n, b, \alpha_k, \Delta_0, L, \sigma^2)$ and $\mathcal{N}_\epsilon (n, b, \alpha_k, \Delta_0, L, \sigma^2)$ are defined as in (\ref{K_N}), the critical batch sizes in Theorem \ref{theorem:3} are used to compute $\mathcal{K}_\epsilon (n, b^\star, \alpha_k, \Delta_0, L, \sigma^2)$ and $\mathcal{N}_\epsilon (n, b^\star, \alpha_k, \Delta_0, L, \sigma^2)$. In {\bf{(Decay 2)}}, we assume that $b^\star = \frac{D_2+1}{\epsilon^2}$. (see also (\ref{cbs})).
\end{theorem}

Theorem \ref{theorem:4} indicates that the iteration and SFO complexities for \rm{\bf{(Constant)}} are smaller than those for \rm{\bf{(Decay 1)}}--\rm{\bf{(Decay 3)}}.

\section{Numerical Results}
\label{sec:4}
We numerically verified the number of iterations and SFO complexities needed to achieve high test accuracy for different batch sizes in training ResNet \citep{he2016deep} and Wide-ResNet \citep{zagoruyko2016wide}. The parameter $\alpha$ used in {\bf (Constant)} was determined by conducting a grid search of $\{0.001, 0.005, 0.01, 0.05, 0.1, 0.5\}$. The parameters $\alpha$ and $T$ used in the decaying learning rates {\bf (Decay 1)}--{\bf (Decay 3)} defined by $\alpha_k = \frac{\alpha}{\left(\left\lfloor\frac{k}{T}\right\rfloor+1\right)^a}$ were determined by a grid search of $\alpha \in \{0.001, 0.1, 0.125, 0.25, 0.5, 1.0\}$ and $T \in \{5, 10, 20, 30, 40, 50\}$. The parameter $a$ was set to $a = \frac{1}{4}$ in {\bf (Decay 1)} and $a = \frac{3}{4}$ in {\bf (Decay 3)}. We compared SGD with SGD with momentum (momentum), Adam, AdamW, and RMSProp. The learning rates and hyperparameters of these four optimizers were determined on the basis of the previous results \citep{adam,loshchilov2018decoupled,rmsprop} (The weight decay used in the momentum was $5 \times 10^{-4}$). The experimental environment consisted of an NVIDIA DGX A100$\times$8GPU and Dual AMD Rome7742 2.25-GHz, 128 Cores$\times$2CPU. The software environment was Python 3.10.6, PyTorch 1.13.1, and CUDA 11.6. The code is available at
\url{https://github.com/imakn0907/SGD_using_decaying}.
 
\subsection{Training ResNet-18 on the CIFAR-10 and CIFAR-100 datasets}
\begin{figure*}[htbp]
\begin{tabular}{cc}
\begin{minipage}[t]{0.5\hsize}
\centering
\includegraphics[width=1\textwidth]{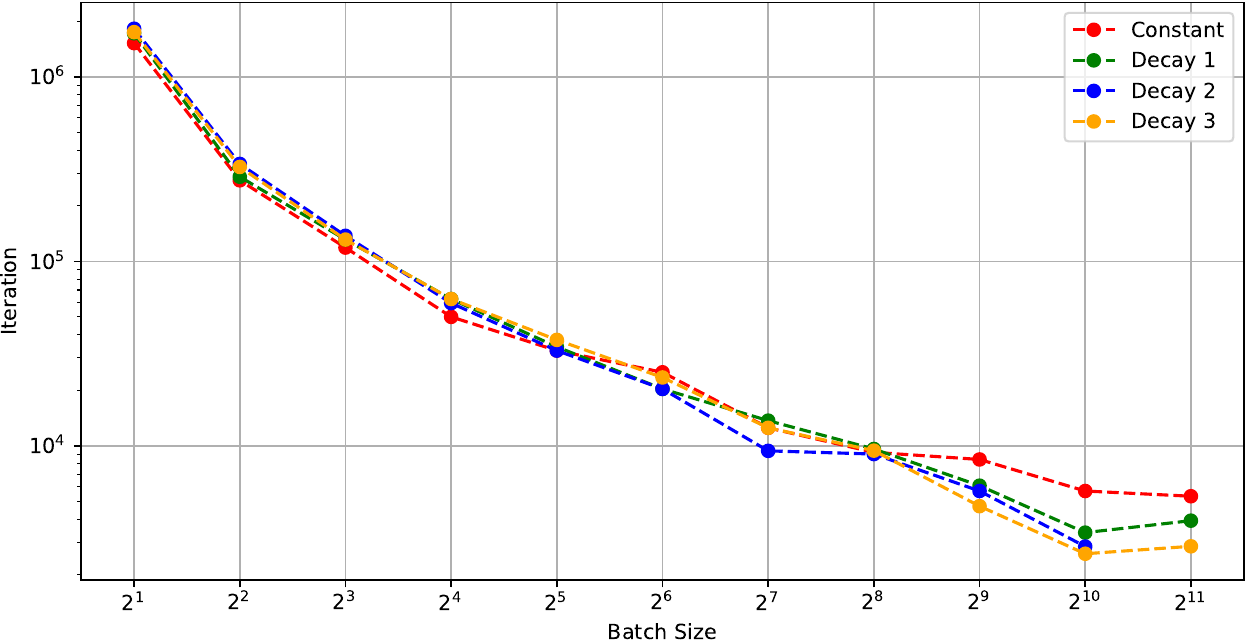}
\caption{Number of iterations needed for SGD with (Constant), (Decay 1), (Decay 2), and (Decay 3) to achieve a test accuracy of $0.9$ versus batch size (ResNet-18 on CIFAR-10)}
\label{fig1}
\end{minipage} &
\begin{minipage}[t]{0.5\hsize}
\centering
\includegraphics[width=\textwidth]{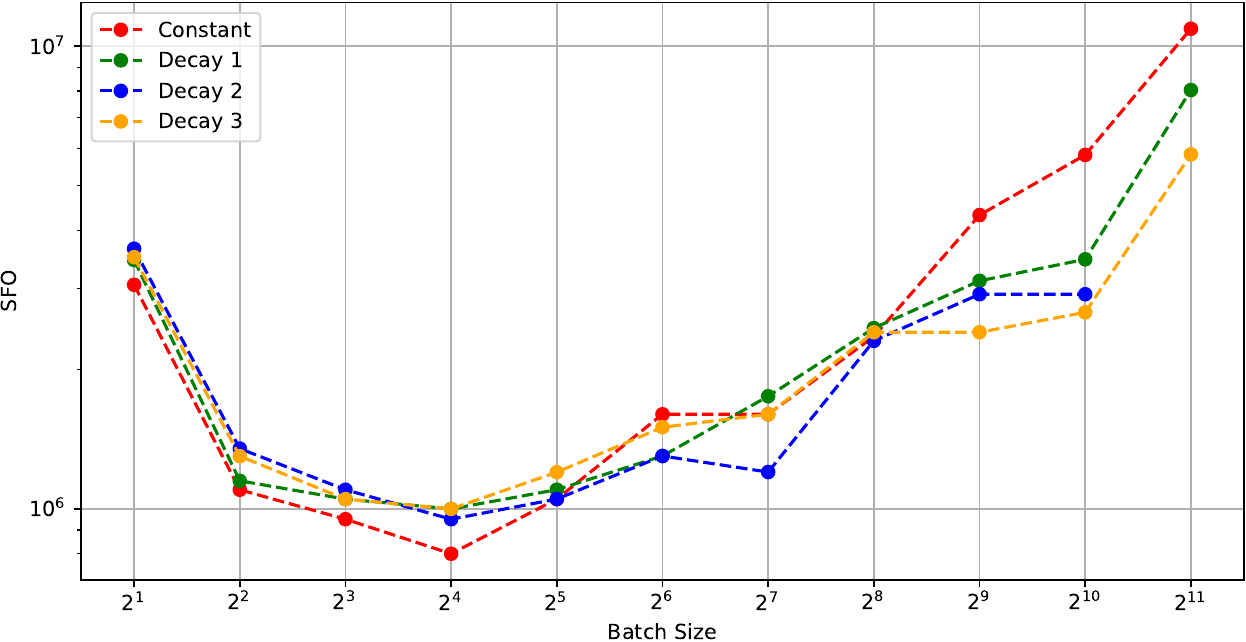}
\caption{SFO complexity needed for SGD with (Constant), (Decay 1), (Decay 2), and (Decay 3) to achieve a test accuracy of $0.9$ versus batch size (ResNet-18 on CIFAR-10)}
\label{fig2}
\end{minipage}
\end{tabular}
\end{figure*}

First, we trained ResNet-18 on the CIFAR-10 dataset. The stopping condition of the optimizers was $200$ epochs. Figure \ref{fig1} indicates that the number of iterations is monotone decreasing and convex with respect to batch size for SGDs using a constant learning rate or a decaying learning rate. Figure \ref{fig2} indicates that, in each case of SGD with {\bf (Constant)}--{\bf(Decay 3)}, a critical batch size $b^\star = 2^4$ exists at which the SFO complexity is minimized. 

\begin{figure*}[htbp]
\begin{tabular}{cc}
\begin{minipage}[t]{0.5\hsize}
\centering
\includegraphics[width=1\textwidth]{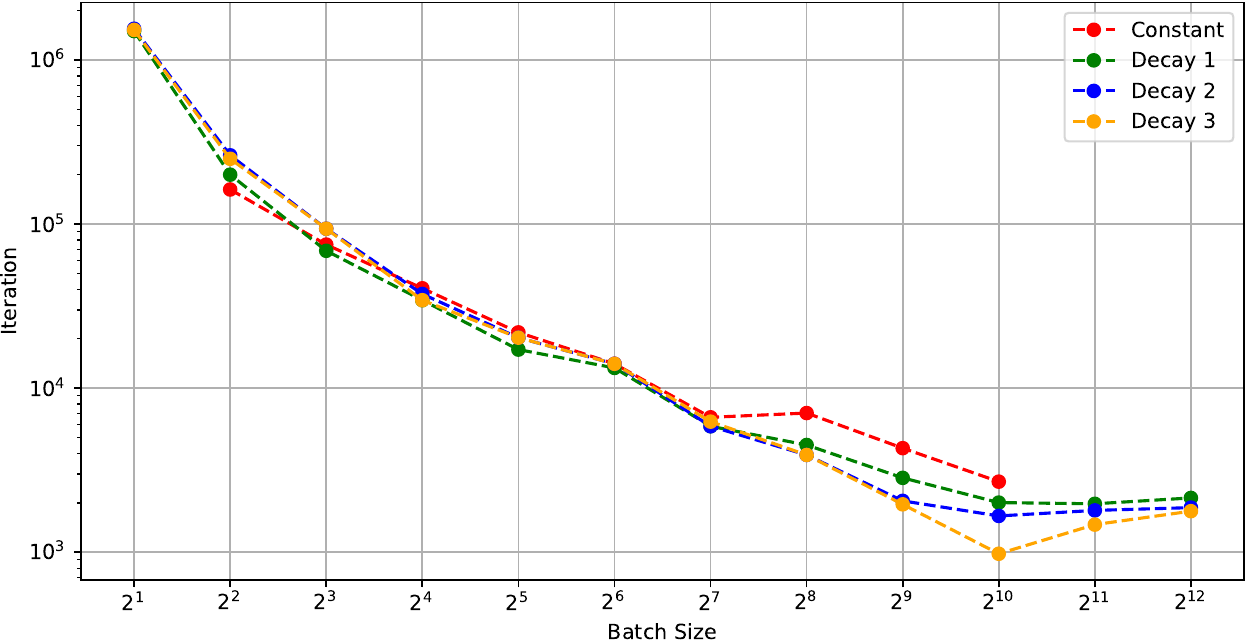}
\caption{Number of iterations needed for SGD with (Constant), (Decay 1), (Decay 2), and (Decay 3) to achieve a test accuracy of $0.6$ versus batch size (ResNet-18 on CIFAR-100)}
\label{fig1_1}
\end{minipage} &
\begin{minipage}[t]{0.5\hsize}
\centering
\includegraphics[width=1\textwidth]{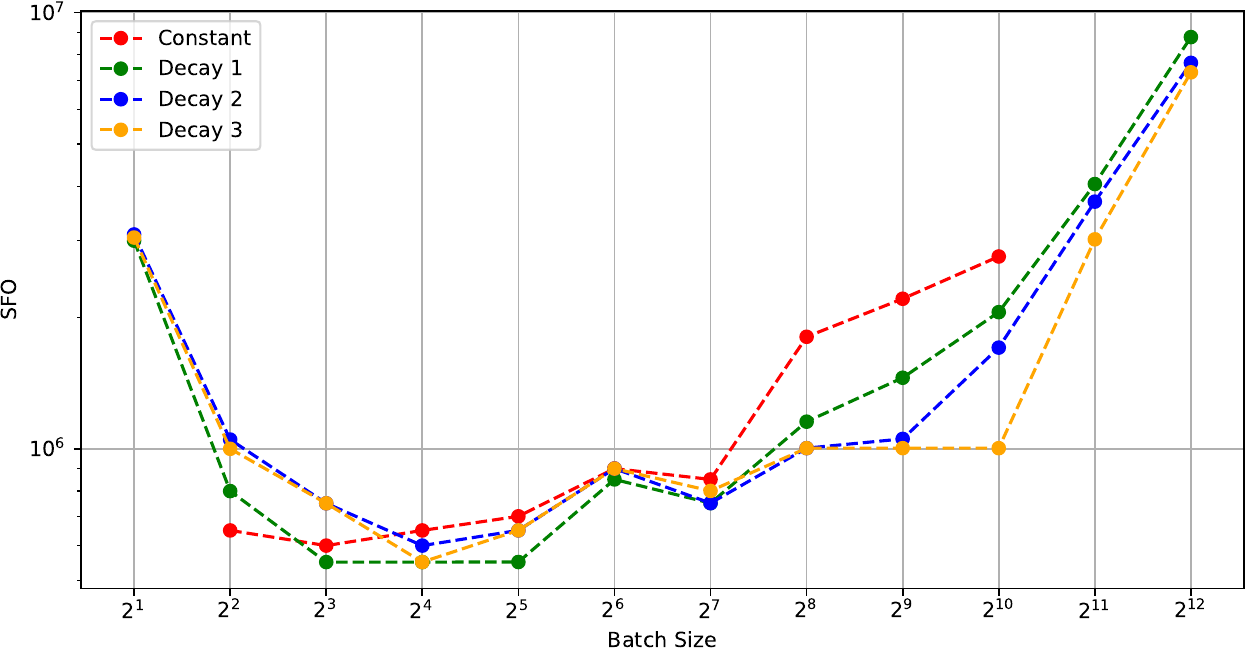}
\caption{SFO complexity needed for SGD with (Constant), (Decay 1), (Decay 2), and (Decay 3) to achieve a test accuracy of $0.6$ versus batch size (ResNet-18 on CIFAR-100)}
\label{fig2_1}
\end{minipage}
\end{tabular}
\end{figure*}

Figures \ref{fig1_1} and \ref{fig2_1} indicate that the number of iterations and the SFO complexity for four different learning rates in achieving a test accuracy of 0.6 when training ResNet-18 on the CIFAR-100 dataset. The figures indicate that critical batch sizes existed when using {\bf (Constant)}--{\bf (Decay 3)}.

\begin{figure*}[htbp]
\begin{tabular}{cc}
\begin{minipage}[t]{0.5\hsize}
\centering
\includegraphics[width=1\textwidth]{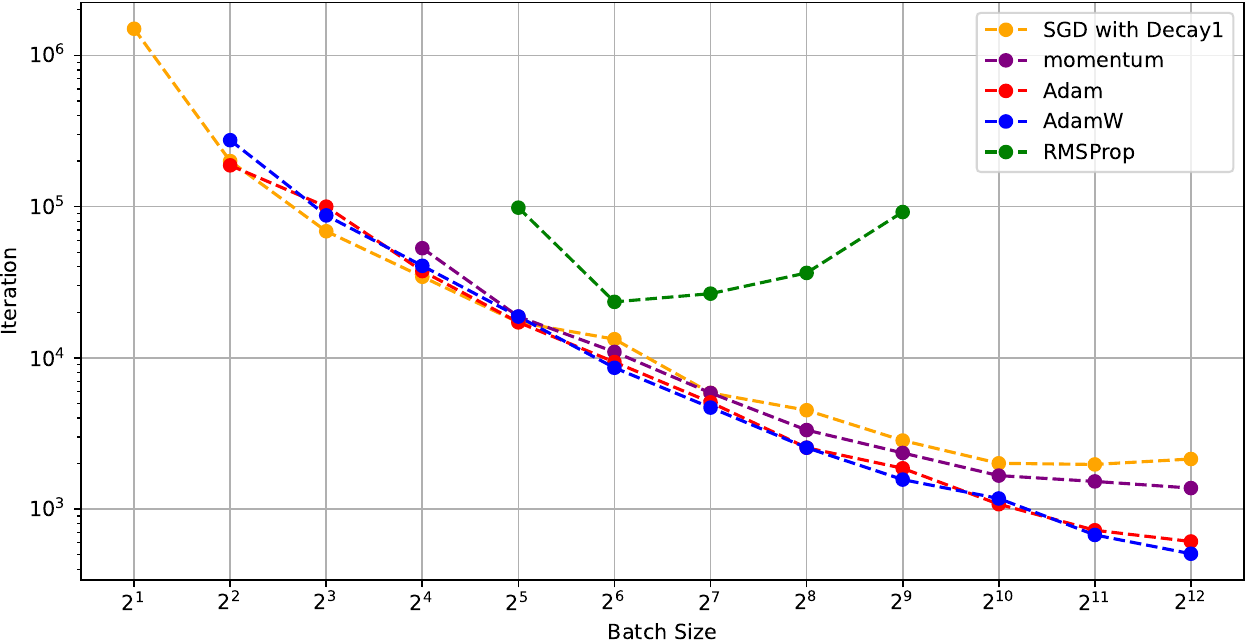}
\caption{Number of iterations needed for SGD with (Decay 1), momentum, Adam, AdamW, and RMSProp to achieve a test accuracy of $0.6$ versus batch size (ResNet-18 on CIFAR-100)}
\label{fig9}
\end{minipage} &
\begin{minipage}[t]{0.5\hsize}
\centering
\includegraphics[width=1\textwidth]{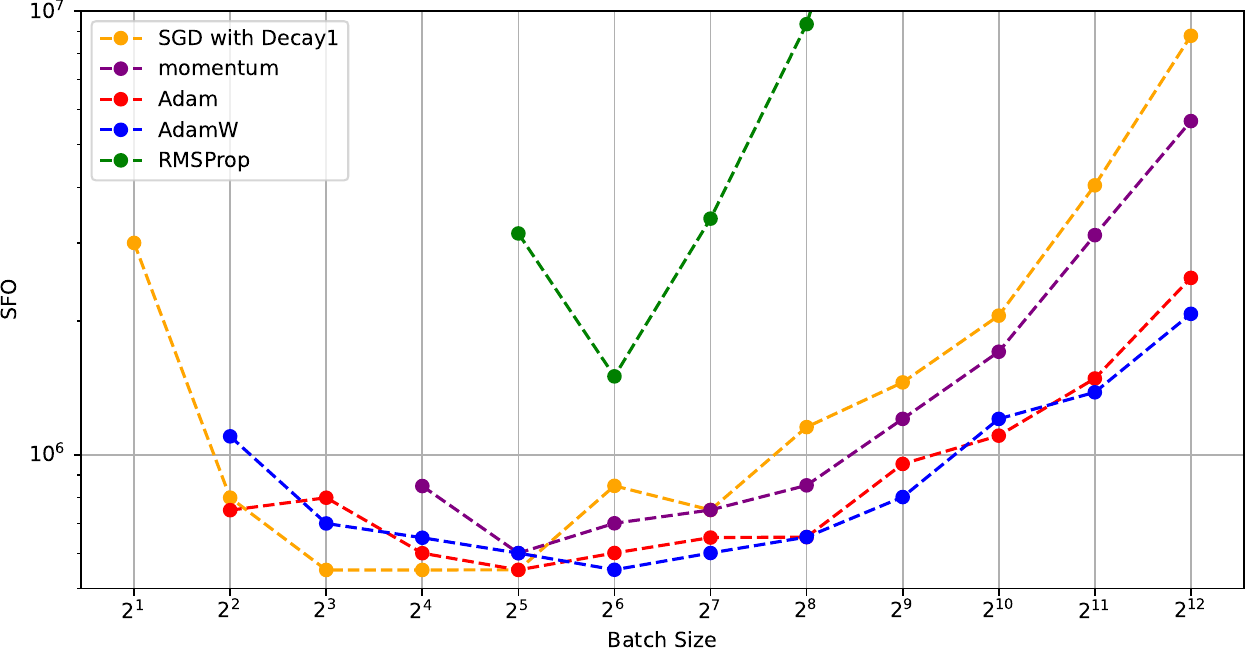}
\caption{SFO complexity needed for SGD with (Decay 1), momentum, Adam, AdamW, and RMSProp to achieve a test accuracy of $0.6$ versus batch size (ResNet-18 on CIFAR-100)}
\label{fig10}
\end{minipage}
\end{tabular}
\end{figure*}

Figures \ref{fig9} and \ref{fig10} compare SGD with {\bf (Decay 1)} with the other optimizers in training ResNet-18 on the CIFAR-100 dataset. These figures indicates that SGD with {\bf (Decay 1)} and a critical batch size ($b = 2^4$) outperformed the other optimizers in the sense of minimizing the number of iterations and the SFO complexity. Figure \ref{fig10} also indicates that the existing optimizers using constant learning rates had critical batch sizes minimizing the SFO complexities.
In particular, AdamW using the critical batch size $b^\star = 2^5$ performed well.

\subsection{Training Wide-ResNet on the CIFAR-10 and CIFAR-100 datasets}
\begin{figure*}[htbp]
\begin{tabular}{cc}
\begin{minipage}[t]{0.5\hsize}
\centering
\includegraphics[width=1\textwidth]{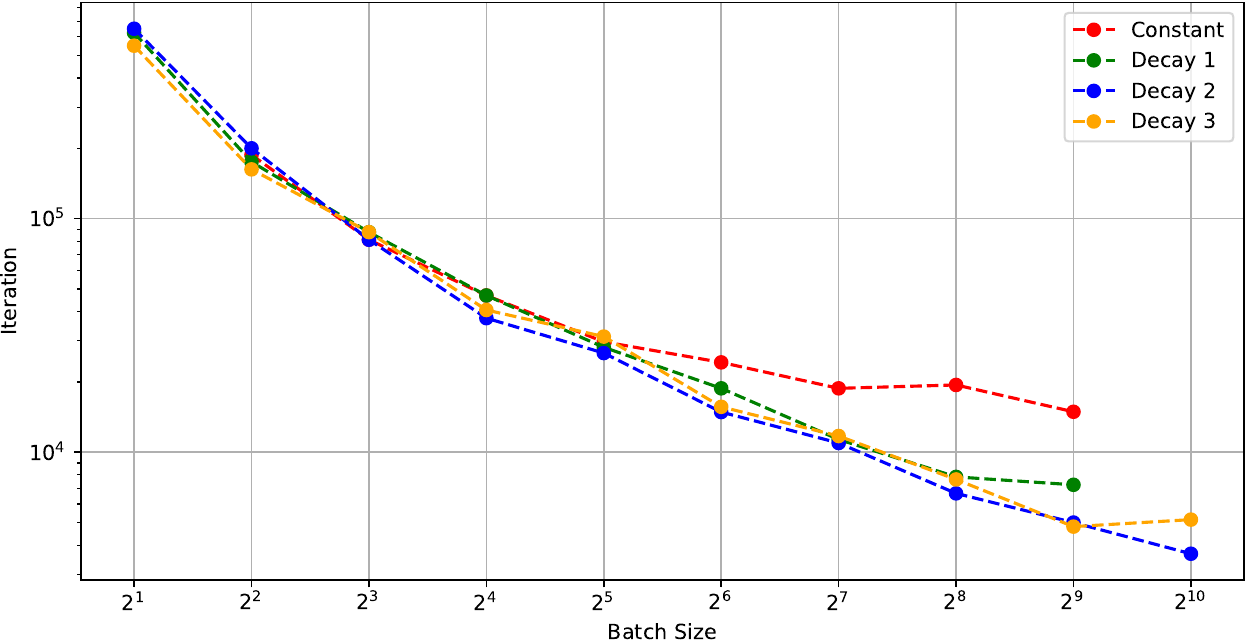}
\caption{Number of iterations needed for SGD with (Constant), (Decay 1), (Decay 2), and (Decay 3) to achieve a test accuracy of $0.9$ versus batch size (Wide-ResNet-28-10 on CIFAR-10)}
\label{fig1_10}
\end{minipage} &
\begin{minipage}[t]{0.5\hsize}
\centering
\includegraphics[width=1\textwidth]{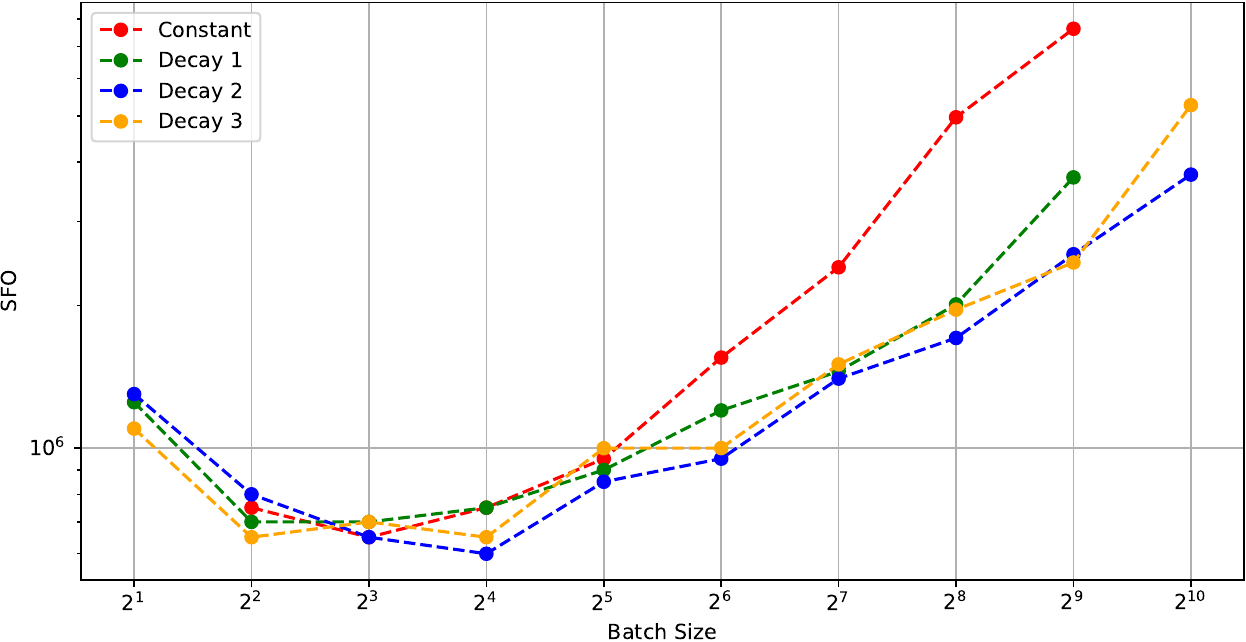}
\caption{SFO complexity needed for SGD with (Constant), (Decay 1), (Decay 2), and (Decay 3) to achieve a test accuracy of $0.9$ versus batch size (Wide-ResNet-28-10 on CIFAR-10)}
\label{fig2_10}
\end{minipage}
\end{tabular}
\end{figure*}

Next, we trained Wide-ResNet-28 \citep{zagoruyko2016wide} on the CIFAR-10 and CIFAR-100 datasets. The stopping condition of the optimizers was $200$ epochs. Figures \ref{fig1_10} and \ref{fig2_10} show that the number of iterations and the SFO complexity of SGD to achieve a test accuracy of $0.9$ (CIFAR-10) versus batch size. Figures \ref{fig1_10} and \ref{fig2_10} indicate that the critical batch size was $b^\star = 2^4$ in each case of SGD using {\bf (Constant)}--{\bf(Decay 3)}. 

\begin{figure*}[htbp]
\begin{tabular}{cc}
\begin{minipage}[t]{0.5\hsize}
\centering
\includegraphics[width=1\textwidth]{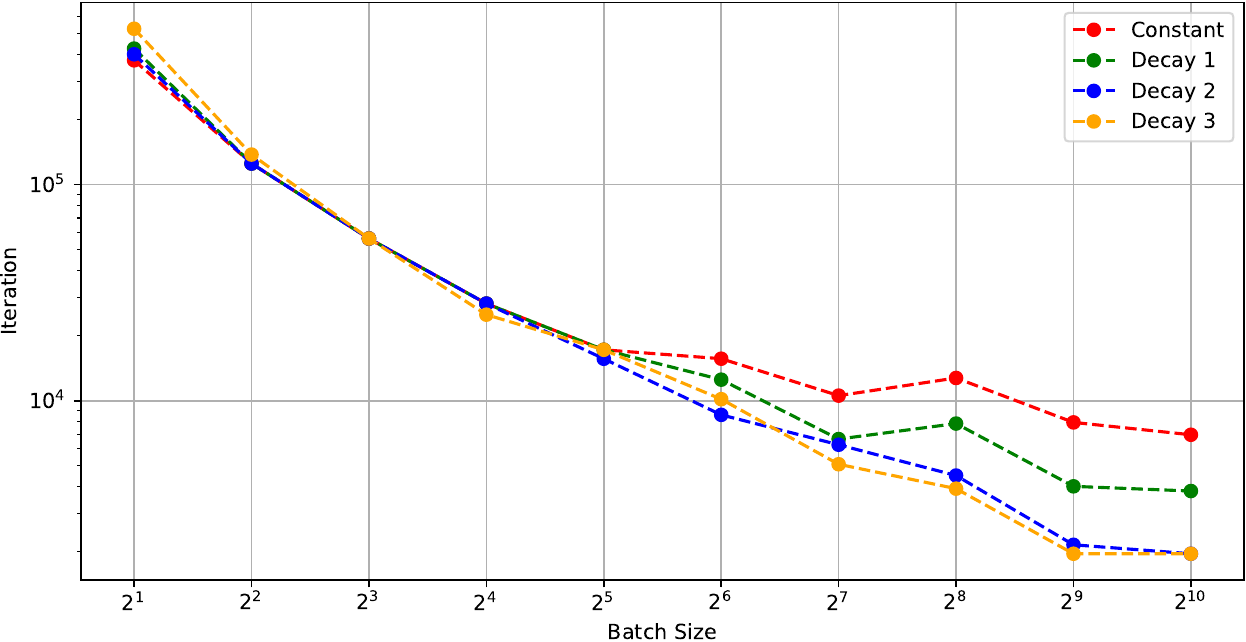}
\caption{Number of iterations needed for SGD with (Constant), (Decay 1), (Decay 2), and (Decay 3) to achieve a test accuracy of $0.6$ versus batch size (WideResNet-28-12 on CIFAR-100)}
\label{fig7}
\end{minipage}
\begin{minipage}[t]{0.5\hsize}
\centering
\includegraphics[width=1\textwidth]{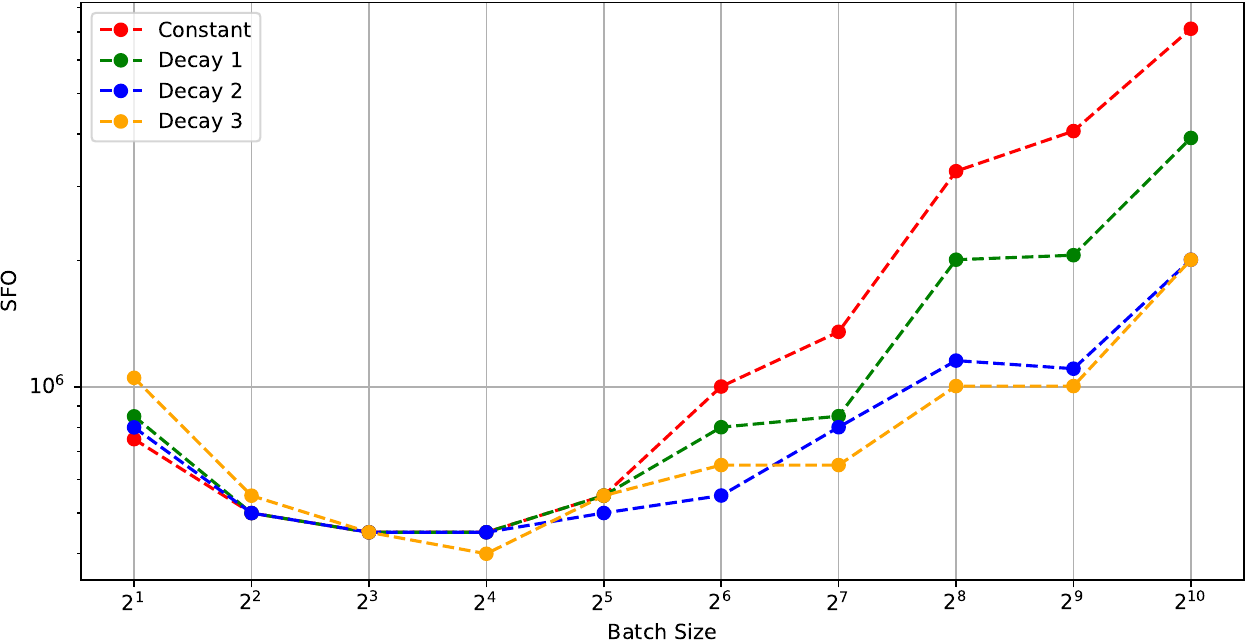}
\caption{SFO complexity needed for SGD with (Constant), (Decay 1), (Decay 2), and (Decay 3) to achieve a test accuracy of $0.6$ versus batch size (WideResNet-28-12 on CIFAR-100)}
\label{fig8}
\end{minipage}
\end{tabular}
\end{figure*}

Figures \ref{fig7} and \ref{fig8} indicate that the number of iterations and the SFO complexity of SGD to achieve a test accuracy of 0.6 (CIFAR-100) versus batch size and show that a critical batch size existed for {\bf (Constant)}--{\bf (Decay 3)}.

\begin{figure*}[htbp]
\begin{tabular}{cc}
\begin{minipage}[t]{0.5\hsize}
\centering
\includegraphics[width=1\textwidth]{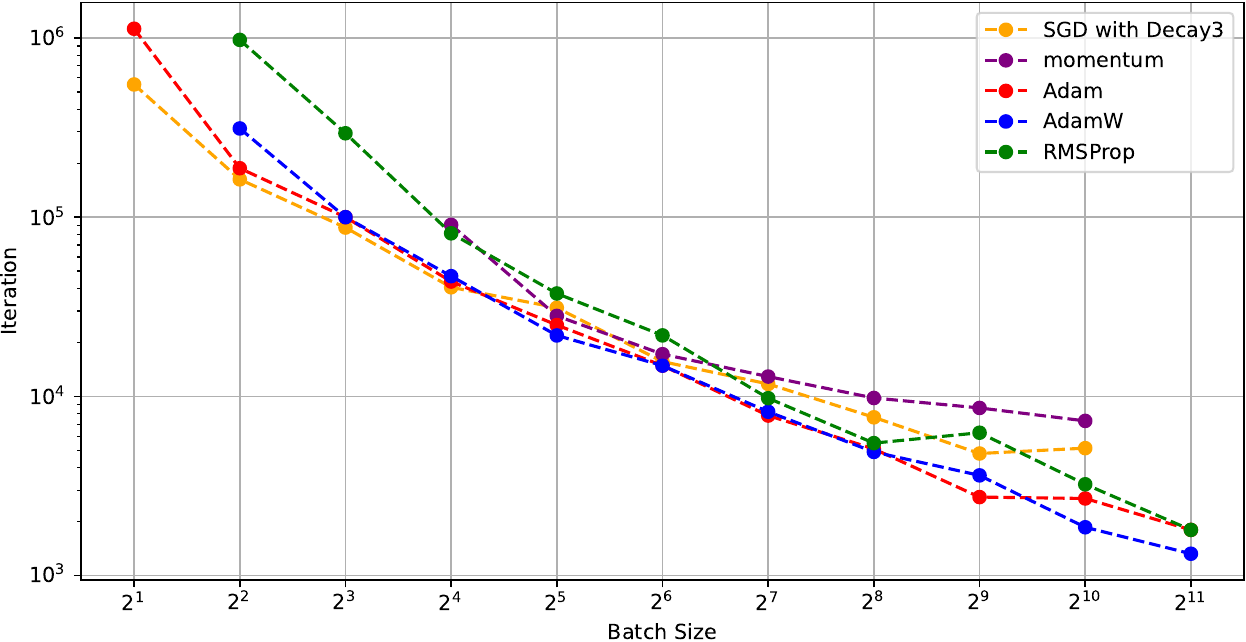}
\caption{Number of iterations needed for SGD with (Decay 3), momentum, Adam, AdamW, and RMSProp to achieve a test accuracy of $0.9$ versus batch size (WideResNet-28-10 on CIFAR-10)}
\label{fig3}
\end{minipage} &
\begin{minipage}[t]{0.5\hsize}
\centering
\includegraphics[width=1\textwidth]{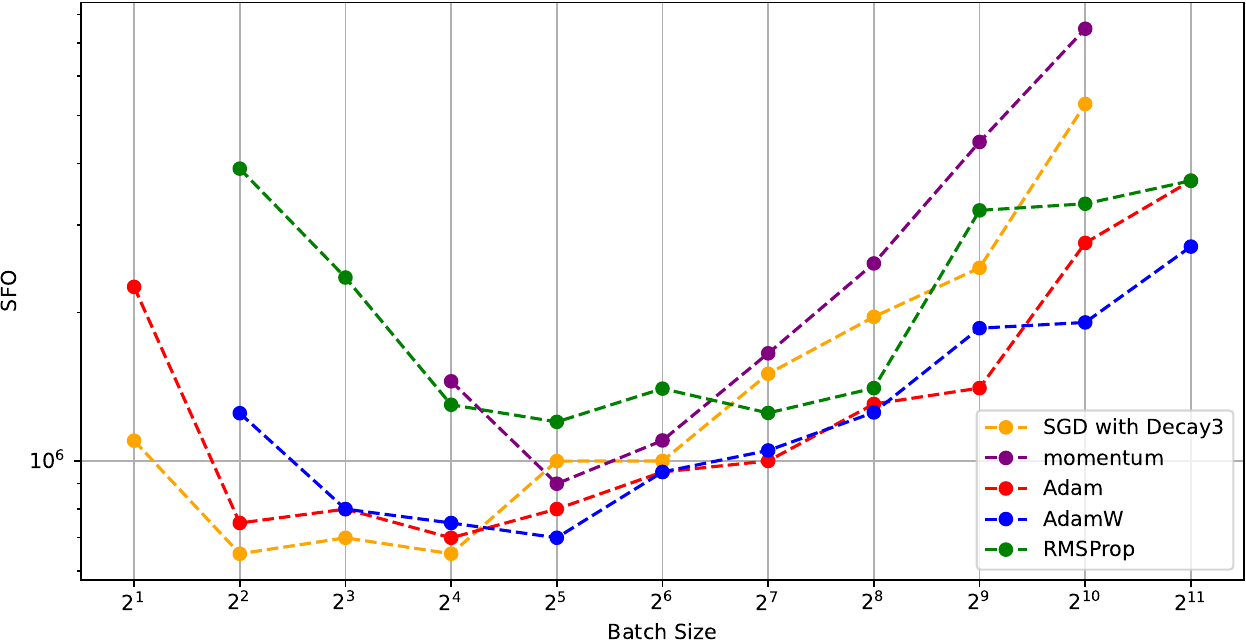}
\caption{SFO complexity needed for SGD with (Decay 3), momentum, Adam, AdamW, and RMSProp to achieve a test accuracy of $0.9$ versus batch size (WideResNet-28-10 on CIFAR-10)}
\label{fig4}
\end{minipage}
\end{tabular}
\end{figure*}

As in Figures \ref{fig9} and \ref{fig10}, Figures \ref{fig3} and \ref{fig4} indicate that SGD using {\bf (Decay 3)} and the existing optimizers using constant learning rates had critical batch sizes minimizing the SFO complexities.

\subsection{Estimation of critical batch sizes}\label{sec:4.3}

\begin{table}[ht]
\centering
\caption{Measured (left) and estimated (right; bold) critical batch sizes (D1 and D3 stand for (Decay 1) and (Decay 3))}
\label{table_measured}
\begin{tabular}{lllll}
\toprule
\multirow{2}{*}{} & \multicolumn{2}{l}{ResNet-18} & \multicolumn{2}{l}{Wide-ResNet-28} \\
\cmidrule{2-5}
 & CIFAR10 & CIFAR100 & CIFAR10 & CIFAR100 \\
\midrule
D1 & $2^4$ & $2^4$ & $2^2$ & $2^3, 2^4$ \\
D3 & $2^4$ \text{ } $\textbf{24}$ & $2^4$ \text{ } $\textbf{24}$ & $2^2, 2^4$ \text{ } $\textbf{6}$ & $2^4$ \text{ } $\textbf{12, 24}$ \\
\bottomrule
\end{tabular}
\end{table}

We estimated the critical batch sizes of {\bf (Decay 3)} using Theorem \ref{theorem:3} and measured the critical batch sizes of {\bf (Decay 1)}. From (\ref{cbs}) and $a = \frac{1}{4}$ ({\bf (Decay 1)}), we have that, for training ResNet-18 on the CIFAR-10 dataset, $b^\star = 2^4 = \frac{(1-a)D_2}{a(1-2a)D_1}$, i.e., $\frac{D_2}{D_1} = \frac{8}{3}$. Then, the estimated critical batch size of SGD using {\bf (Decay 3)} ($a = \frac{3}{4}$) for training ResNet-18 on the CIFAR-10 dataset is 
\begin{align*}
b^\star &= \frac{2 a^2}{(1-a)(2a -1)} \frac{D_2}{D_1} = \frac{2 a^2}{(1-a)(2a -1)} \frac{8}{3}\\ 
&= 24 \in (2^4, 2^5),
\end{align*}
which implies that the estimated critical batch size $b^\star = 24$ is close to the measured size $b = 2^4$. We also found that the estimated critical batch sizes are close to the measured critical batch sizes (see Table \ref{table_measured}).

\section{Conclusion and future work}
This paper investigated the required number of iterations and SFO complexities for SGD using constant or decay learning rates to achieve an $\epsilon$--approximation. Our theoretical analyses indicated that the number of iterations needed for an $\epsilon$--approximation is monotone decreasing and convex with respect to the batch size and the SFO complexity needed for an $\epsilon$--approximation is convex with respect to the batch size. Moreover, we showed that SGD using a critical batch size reduces the SFO complexity. The numerical results indicated that SGD using the critical batch size performs better than the existing optimizers in the sense of minimizing the SFO complexity. We also estimated critical batch sizes of SGD using our theoretical results and showed that they are close to the measured critical batch sizes. 

The results in this paper can be only applied to SGD. This is a limitation of our work. Hence, in the future, we should investigate whether our results can be applied to variants of SGD, such as the momentum methods and adaptive methods.

(3669 words)

\bibliographystyle{tfnlm}
\bibliography{main_op}

\newpage
\appendix
\onecolumn
\section{Appendix}
\subsection{Lemma}
First, we will prove the following lemma.

\begin{lemma}\label{lem:1}
The sequence $(\bm{\theta}_k)_{k \in \mathbb{N}}$ generated by Algorithm \ref{algo:1} under (C1)--(C3) satisfies that, for all $K \geq 1$,
\begin{align*}
\sum_{k=0}^{K-1} \alpha_k \left(1 - \frac{L \alpha_k}{2} \right)
\mathbb{E}\left[ \|\nabla f(\bm{\theta}_{k})\|^2 \right] 
\leq
\mathbb{E}\left[f(\bm{\theta}_{0}) - f_\star \right]
+ \frac{L \sigma^2}{2b} \sum_{k=0}^{K-1} \alpha_k^2,
\end{align*}
where $\mathbb{E}$ stands for the total expectation. 
\end{lemma}

{\em Proof:} Condition (C1) ($L$-smoothness of $f$) implies that the descent lemma holds, i.e., for all $k \in \mathbb{N}$, 
\begin{align*}
f(\bm{\theta}_{k+1}) 
\leq f(\bm{\theta}_{k}) + \langle \nabla f(\bm{\theta}_{k}),
\bm{\theta}_{k+1} - \bm{\theta}_{k} \rangle + \frac{L}{2} 
\|\bm{\theta}_{k+1} - \bm{\theta}_{k}\|^2,
\end{align*}
which, together with $\bm{\theta}_{k+1} := \bm{\theta}_{k} - \alpha_k \nabla f_{B_k}(\bm{\theta}_k)$, implies that 
\begin{align}\label{ineq_main}
f(\bm{\theta}_{k+1}) 
\leq f(\bm{\theta}_{k}) - \alpha_k \langle \nabla f(\bm{\theta}_{k}),
\nabla f_{B_k}(\bm{\theta}_k) \rangle + \frac{L \alpha_k^2}{2} 
\|\nabla f_{B_k}(\bm{\theta}_k)\|^2.
\end{align}
Condition (C2) guarantees that 
\begin{align}\label{estimation}
\mathbb{E}_{\xi_k} \left[\nabla f_{B_k} (\bm{\theta}_k) | \bm{\theta}_k \right] 
= \nabla f(\bm{\theta}_k) \text{ and } 
\mathbb{E}_{\xi_k} \left[\| \nabla f_{B_k} (\bm{\theta}_k) - \nabla f(\bm{\theta}_k) \|^2 | \bm{\theta}_k \right] 
\leq \frac{\sigma^2}{b}.
\end{align}
Hence, we have 
\begin{align}
\mathbb{E}_{\xi_k} \left[\|\nabla f_{B_k} (\bm{\theta}_k)\|^2 | \bm{\theta}_k \right]
&= 
\mathbb{E}_{\xi_k} \left[\|\nabla f_{B_k} (\bm{\theta}_k) - \nabla f(\bm{\theta}_k) + \nabla f(\bm{\theta}_k) \|^2 | \bm{\theta}_k \right]\nonumber\\
&=
\mathbb{E}_{\xi_k} \left[\|\nabla f_{B_k} (\bm{\theta}_k) - \nabla f(\bm{\theta}_k) \|^2 | \bm{\theta}_k \right]
+ 2 \mathbb{E}_{\xi_k} \left[ \langle \nabla f_{B_k} (\bm{\theta}_k) - \nabla f(\bm{\theta}_k),
\nabla f(\bm{\theta}_k) \rangle | \bm{\theta}_k \right]\nonumber\\ 
&\quad + 
\mathbb{E}_{\xi_k} \left[\| \nabla f(\bm{\theta}_k) \|^2 | \bm{\theta}_k \right] \nonumber\\
&\leq \frac{\sigma^2}{b} + \mathbb{E}_{\xi_k} \left[\| \nabla f(\bm{\theta}_k) \|^2 \right]. \label{e_2}
\end{align}
Taking the expectation conditioned on $\bm{\theta}_k$ on both sides of (\ref{ineq_main}), together with (\ref{estimation}) and (\ref{e_2}), guarantees that, for all $k \in \mathbb{N}$,
\begin{align*}
\mathbb{E}_{\xi_k} \left[f(\bm{\theta}_{k+1}) | \bm{\theta}_k \right] 
&\leq 
f(\bm{\theta}_{k}) - \alpha_k \mathbb{E}_{\xi_k} \left[ \langle \nabla f(\bm{\theta}_{k}),
\nabla f_{B_k}(\bm{\theta}_k) \rangle | \bm{\theta}_k \right] 
+ \frac{L \alpha_k^2}{2} 
\mathbb{E}_{\xi_k} \left[ \|\nabla f_{B_k}(\bm{\theta}_k)\|^2| \bm{\theta}_k \right]\\
&\leq 
f(\bm{\theta}_{k}) - \alpha_k \|\nabla f(\bm{\theta}_{k})\|^2
+ \frac{L \alpha_k^2}{2} 
\left( \frac{\sigma^2}{b} + \|\nabla f(\bm{\theta}_{k})\|^2 \right).
\end{align*}
Hence, taking the total expectation on both sides of the above inequality ensures that, for all $k \in \mathbb{N}$,
\begin{align*}
\alpha_k \left(1 - \frac{L \alpha_k}{2} \right)
\mathbb{E}\left[ \|\nabla f(\bm{\theta}_{k})\|^2 \right] 
\leq
\mathbb{E}\left[f(\bm{\theta}_{k}) - f(\bm{\theta}_{k+1}) \right]
+ \frac{L \sigma^2 \alpha_k^2}{2b}.
\end{align*}
Let $K \geq 1$. Summing the above inequality from $k = 0$ to $k = K-1$ ensures that
\begin{align*}
\sum_{k=0}^{K-1} \alpha_k \left(1 - \frac{L \alpha_k}{2} \right)
\mathbb{E}\left[ \|\nabla f(\bm{\theta}_{k})\|^2 \right] 
\leq
\mathbb{E}\left[f(\bm{\theta}_{0}) - f(\bm{\theta}_{K}) \right]
+ \frac{L \sigma^2}{2b} \sum_{k=0}^{K-1} \alpha_k^2,
\end{align*}
which, together with (C1) (the lower bound $f_\star$ of $f$), implies that the assertion in Lemma \ref{lem:1} holds.
\qed

\subsection{Proof of Theorem \ref{theorem:1}}
\label{a_1}
{\bf (Constant):} Lemma \ref{lem:1} with $\alpha_k = \alpha$ implies that
\begin{align*}
\alpha \left(1 - \frac{L \alpha}{2} \right)
\sum_{k=0}^{K-1}
\mathbb{E}\left[ \|\nabla f(\bm{\theta}_{k})\|^2 \right] 
\leq
\mathbb{E}\left[f(\bm{\theta}_{0}) - f_\star \right]
+ \frac{L \sigma^2 \alpha^2 K}{2b}.
\end{align*}
Since $\alpha < \frac{2}{L}$, we have that
\begin{align*}
\min_{k\in [0:K-1]} \mathbb{E}\left[ \|\nabla f(\bm{\theta}_{k})\|^2 \right]
\leq
\frac{1}{K} \sum_{k=0}^{K-1}
\mathbb{E}\left[ \|\nabla f(\bm{\theta}_{k})\|^2 \right] 
\leq
\underbrace{\frac{2 (f(\bm{\theta}_{0}) - f_\star )}{(2 - L\alpha) \alpha}}_{C_1} \frac{1}{K}
+ 
\underbrace{\frac{L \sigma^2 \alpha}{2 - L \alpha}}_{C_2} \frac{1}{b}.
\end{align*} 

{\bf (Decay):} Since $(\alpha_k)_{k\in \mathbb{N}}$ converges to $0$, there exists $k_0 \in \mathbb{N}$ such that, for all $k \geq k_0$, $\alpha_k < \frac{2}{L}$. We assume that $k_0 = 0$ (see Section \ref{subsec:2.2.3}). Lemma \ref{lem:1} ensures that, for all $K \geq 1$,
\begin{align*}
\sum_{k=0}^{K-1} \alpha_k \left(1 - \frac{L \alpha_k}{2} \right)
\mathbb{E}\left[ \|\nabla f(\bm{\theta}_{k})\|^2 \right]
\leq
\mathbb{E}\left[f(\bm{\theta}_{0}) - f_\star \right]
+ \frac{L \sigma^2}{2b} \sum_{k=0}^{K-1} \alpha_k^2,
\end{align*}
which, together with $\alpha_{k+1} \leq \alpha_k < \frac{2}{L}$ $(k\in \mathbb{N})$, implies that 
\begin{align*}
\alpha_{K-1} \left(1 - \frac{L \alpha_{0}}{2} \right) \sum_{k=0}^{K-1} \mathbb{E}\left[ \|\nabla f(\bm{\theta}_{k})\|^2 \right]
\leq
\mathbb{E}\left[f(\bm{\theta}_{0}) - f_\star \right]
+ \frac{L \sigma^2}{2b} \sum_{k=0}^{K-1} \alpha_k^2.
\end{align*}
Hence, we have that 
\begin{align*}
\sum_{k=0}^{K-1} \mathbb{E}\left[ \|\nabla f(\bm{\theta}_{k})\|^2 \right]
\leq \frac{2 ( f(\bm{\theta}_{0}) - f_\star) }{(2 - L \alpha_{0}) \alpha_{K-1}}
+ 
\frac{L \sigma^2}{b(2 - L \alpha_{0}) \alpha_{K-1}} \sum_{k=0}^{K-1} \alpha_k^2,
\end{align*}
which implies that 
\begin{align*}
\min_{k \in [0:K-1]} \mathbb{E}\left[ \|\nabla f(\bm{\theta}_{k})\|^2 \right]
\leq
\frac{2 ( f(\bm{\theta}_{0}) - f_\star) }{2 - L \alpha_{0}}
\frac{1}{K \alpha_{K-1}}
+ 
\frac{1}{b} \frac{L \sigma^2}{2 - L \alpha_{0}} \frac{1}{K \alpha_{K-1}} \sum_{k=0}^{K-1} \alpha_k^2.
\end{align*}
Meanwhile, we have that 
\begin{align*}
\sum_{k=0}^{K-1} \alpha_k^2 
\leq 
\sum_{k=0}^{K-1} \frac{T\alpha^2}{(k+1)^{2a}}
\leq 
T\alpha^2 \left(
1 + \int_0^{K-1} \frac{\mathrm{d}t}{(t+1)^{2a}}
\right)
\leq 
\begin{cases}
\displaystyle{\frac{T\alpha^2}{1 - 2a} K^{1 - 2a}} &\textbf{ (Decay 1)}\\
\displaystyle{T\alpha^2(1 + \log K)} &\textbf{ (Decay 2)}\\
\displaystyle{\frac{2a T\alpha^2}{2a -1}} &\textbf{ (Decay 3)}
\end{cases}
\end{align*}
and 
\begin{align*}
\alpha_{K-1} = \frac{\alpha}{\left(\lfloor\frac{K-1}{T}\rfloor + 1\right)^a} \geq \frac{\alpha}{\left(\frac{K-1}{T} + 1\right)^a} \geq \frac{\alpha}{K^a}. 
\end{align*}
Here, we define 
\begin{align*}
D_1 := 
\displaystyle{\frac{2\left(f(\bm{\theta}_0)-f_\star\right)}{\alpha(2-L\alpha)}}
\text{ and }
D_2 := \frac{T\alpha^2L\sigma^2}{2-L\alpha}.
\end{align*}
Accordingly, we have that 
\begin{align*}
\min_{k \in [0:K-1]} \mathbb{E}\left[ \|\nabla f(\bm{\theta}_{k})\|^2 \right]
\leq
\begin{cases}
\displaystyle{\frac{D_1}{K^{1-a}} + \frac{D_2}{(1-2a) K^a b}} &\textbf{ (Decay 1)}\\
\displaystyle{\frac{D_1}{\sqrt{K}} + \frac{D_2 (1 + \log K)}{\sqrt{K} b}}&\textbf{ (Decay 2)}\\
\displaystyle{\frac{D_1}{K^{1-a}} + \frac{2 a D_2}{(2a -1) K^{1-a} b}} &\textbf{ (Decay 3)}
\end{cases}
\end{align*}
which, together with $\log K < \sqrt{K}$ and the condition on $a$, implies that 
\begin{align*}
\min_{k \in [0:K-1]} \mathbb{E}\left[ \|\nabla f(\bm{\theta}_{k})\|^2 \right]
\leq
\begin{cases}
\displaystyle{\frac{D_1}{K^{a}} + \frac{D_2}{(1-2a) K^a b}} &\textbf{ (Decay 1)}\\
\displaystyle{\frac{D_1}{\sqrt{K}} + \left(\frac{1}{\sqrt{K}} + 1 \right) \frac{D_2}{b}}&\textbf{ (Decay 2)}\\
\displaystyle{\frac{D_1}{K^{1-a}} + \frac{2 a D_2}{(2a -1) K^{1-a} b}} &\textbf{ (Decay 3)}.
\end{cases}
\end{align*}
\qed

\subsection{Proof of Theorem \ref{theorem:2}}
\label{a_2}
(i) Let us consider the case of \textbf{(Constant)}. We consider that the upper bound $\frac{C_1}{K} + \frac{C_2}{b}$ in Theorem \ref{theorem:1} is equal to $\epsilon^2$. This implies that $K = \frac{C_1 b}{\epsilon^2 b - C_2}$ achieves an $\epsilon$--approximation. A discussion similar to the one showing that $K = \frac{C_1 b}{\epsilon^2 b - C_2}$ is an $\epsilon$--approximation ensures that the assertion in Theorem \ref{theorem:2}(i) is true.

(ii) It is sufficient to prove that $K' = K'(b) < 0$ and $K'' = K''(b) > 0$ hold. 

{\bf (Constant):} Let $K = \frac{C_1 b}{\epsilon^2 b - C_2}$. Then, we have that
\begin{align*} 
	K' &= \frac{C_1(\epsilon^2b - C_2) - \epsilon^2C_1b}{(\epsilon^2b - C_2)^2} 
	 = -\frac{C_1C_2}{(\epsilon^2b - C_2)^2} < 0, \\
	K'' &= \frac{2\epsilon^2C_1C_2(\epsilon^2b - C_2)}{(\epsilon^2b - C_2)^4} 
	 = \frac{2\epsilon^2C_1C_2((\frac{C_1}{K} + \frac{C_2}{b})b - C_2)}{(\epsilon^2b - C_2)^4} 
	 = \frac{2\epsilon^2C_1^2C_2^2}{K(\epsilon^2b - C_2)^4} > 0.
\end{align*}

{\bf (Decay 1):} Let $K = (\frac{1}{\epsilon^2}(D_1 + \frac{D_2}{(1 - 2a)b}))^{\frac{1}{a}}$. Then, we have that 
\begin{align*}
	K' &= \frac{1}{a} \left\{\frac{1}{\epsilon^2} \left(D_1 + \frac{D_2}{(1 - 2a)b} \right) \right\}^{\frac{1}{a} - 1} \left(-\frac{D_2}{\epsilon^2(1 - 2a)b^2} \right)\\ 
	 &= -\frac{D_2}{a\epsilon^2(1 - 2a)b^2} \left\{ \frac{1}{\epsilon^2} \left(D_1 + \frac{D_2}{(1 - 2a)b}\right) \right\}^{\frac{1 - a}{a}} < 0, \\
	K'' &= \frac{2D_2}{a\epsilon^2(1 - 2a)b^3}\left\{\frac{1}{\epsilon^2} \left(D_1 + \frac{D_2}{(1 - 2a)b} \right)\right\}^{\frac{1 - a}{a}}\\
	 &\quad + \frac{2(1 - a)D_2}{a^2\epsilon^2(1 - 2a)b^3}\left\{\frac{1}{\epsilon^2}\left(D_1 + \frac{D_2}{(1 - 2a)b}\right)\right\}^{\frac{1 - a}{a} - 1}\frac{D_2}{\epsilon^2(1 - 2a)b^2} > 0.
\end{align*}

{\bf (Decay 2):} Let $K = (\frac{bD_1 + D_2}{b\epsilon^2 - D_2})^2$. Then, we have that
\begin{align*}
K' &= \frac{2D_1(bD_1 + D_2)(b\epsilon^2 - D_2)^2 - 2\epsilon^2(b\epsilon^2 - D_2)(bD_1 + D_2)^2}{(b\epsilon^2 - D_2)^4},
\end{align*}
which, together with $b\epsilon^2 - D_2 > 0 $ and $D_1 > \epsilon^2$, implies that
\begin{align*}
(b \epsilon^2 - D_2)^3 K' &= 2 D_1(bD_1 + D_2)(b \epsilon^2 - D_2) - 2 \epsilon^2 (b D_1 + D_2)^2\\
&= 2(bD_1+D_2)\left\{ D_1(b\epsilon^2-D_2)-\epsilon^2(bD_1+D_2) \right\} \\
&= -2(bD_1+D_2)(D_1D_2-\epsilon^2D_2)\\
&= -2D_2(bD_1+D_2)(D_1-\epsilon^2) < 0.
\end{align*}
Moreover,
\begin{align*}
K'' &= \frac{-2D_1D_2(D_1 - \epsilon^2)(b\epsilon^2 - D_2)^3 +
6D_2\epsilon^2(b\epsilon^2 - D_2)^2(bD_1+D_2)(D_1-\epsilon^2)}{(b\epsilon^2 - D_2)^6},
\end{align*}
which implies that
\begin{align*}
(b\epsilon^2 - D_2)^4K'' &= -2D_1D_2(D_1 - \epsilon^2)(b\epsilon^2 - D_2) + 6D_2\epsilon^2(bD_1+D_2)(D_1-\epsilon^2) \\
&= 2D_2(D_1-\epsilon^2)\left\{ -D_1(b\epsilon^2-D_2)+3\epsilon^2(bD_1+D_2)\right\} \\
&= 2D_2(D_1-\epsilon^2)(2D_1\epsilon^2b+D_1D_2+3D_2\epsilon^2) > 0.
\end{align*}

{\bf (Decay 3):} Let $K = (\frac{1}{\epsilon^2}(D_1 + \frac{2aD_2}{(2a - 1)b}))^{\frac{1}{1 - a}}$. Then, we have that
\begin{align*}
	K' &= \frac{1}{1 - a}\left\{\frac{1}{\epsilon^2}\left(D_1 + \frac{2aD_2}{(2a - 1)b} \right)\right\}^{\frac{1}{1 - a} - 1}\left(-\frac{2aD_2}{\epsilon^2(2a - 1)b^2} \right) \\
	 &= -\frac{2aD_2}{\epsilon^2(1 - a)(2a - 1)b^2}\left\{\frac{1}{\epsilon^2}\left(D_1 + \frac{2aD_2}{(2a - 1)b}\right)\right\}^{\frac{a}{1 - a}} < 0, \\
	K'' &= \frac{4aD_2}{\epsilon^2(1 - a)(2a - 1)b^3}\left\{ \frac{1}{\epsilon^2}\left(D_1 + \frac{2aD_2}{(2a - 1)b}\right)\right\}^{\frac{a}{1 - a}} \\
	 &\quad + \frac{2a^2D_2}{\epsilon^2(1 - a)^2(2a - 1)b^2}\left\{\frac{1}{\epsilon^2}\left(D_1 + \frac{2aD_2}{(2a - 1)b}\right)\right\}^{\frac{a}{1 - a} - 1} \frac{2aD_2}{\epsilon^2(2a - 1)b^2} > 0.
\end{align*}
\qed

\subsection{Proof of Theorem \ref{theorem:3}}
\label{a_3}
{\bf (Constant):} Let $N = \frac{C_1b^2}{\epsilon^2b - C_2}$. Then, we have that
\begin{align*}
	N' &= \frac{2C_1b(\epsilon^2b - C_2) - \epsilon^2C_1b^2}{(\epsilon^2b - C_2)^2} = \frac{C_1b(\epsilon^2b - 2C_2)}{(\epsilon^2b - C_2)^2}. 
\end{align*}
If $N' = 0$, we have that $\epsilon^2b - 2C_2 = 0$, i.e., $b = \frac{2C_2}{\epsilon^2}$. Moreover,
\begin{align*}
	N'' &= \frac{(2\epsilon^2C_1b - 2C_1C_2)(\epsilon^2b - C_2)^2 - 2\epsilon^2(\epsilon^2b - C_2)(\epsilon^2C_1b^2 - 2C_1C_2b)}{(\epsilon^2b - C_2)^4} \\
	(\epsilon^2b - C_2)^3N'' &= (2\epsilon^2C_1b - 2C_1C_2)(\epsilon^2b - C_2) - 2\epsilon^2(\epsilon^2C_1b^2 - 2C_1C_2b) \\
	 &= 2C_1C_2^2 > 0,
\end{align*}
which implies that $N$ is convex. Hence, there is a critical batch size $b^{\star} = \frac{2C_2}{\epsilon^2} > 0$ at which $N$ is minimized.

{\bf (Decay 1):} Let $N = Kb$. Then, we have that
\begin{align*}
	N' &= K + bK' \\
	 &= \left\{\frac{1}{\epsilon^2}\left(D_1 + \frac{D_2}{(1 - 2a)b} \right)\right\}^\frac{1}{a}
	 + \frac{1}{a}\left\{ \frac{1}{\epsilon^2}\left(D_1 + \frac{D_2}{(1 - 2a)b} \right)\right\}^{\frac{1}{a} - 1} \left(-\frac{D_2}{\epsilon^2(1 - 2a)b^2} \right)b \\
	 &= \left\{ \frac{1}{\epsilon^2} \left(D_1 + \frac{D_2}{(1 - 2a)b} \right)\right\}^{\frac{1}{a} - 1}\left\{\frac{1}{\epsilon^2}\left(D_1 + \frac{D_2}{(1 - 2a)b} \right) -\frac{D_2}{a\epsilon^2(1 - 2a)b} \right\}.
\end{align*}
If $N' = 0$, we have that 
\begin{align*}
	\frac{1}{\epsilon^2}\left( D_1 + \frac{D_2}{(1 - 2a)b} \right) -\frac{D_2}{a\epsilon^2(1 - 2a)b} = 0, \text{ i.e., }
	b = \frac{D_2(a - 1)}{aD_1(2a - 1)}.
\end{align*}
Moreover, 
\begin{align*}
	N'' &= K' + (K' + bK'') = 2K' + bK'' \\
	 &= -\frac{2D_2}{a\epsilon^2(1 - 2a)b^2}\left\{\frac{1}{\epsilon^2}\left(D_1 + \frac{D_2}{(1 - 2a)b} \right)\right\}^{\frac{1 - a}{a}} + \frac{2D_2}{a\epsilon^2(1 - 2a)b^2}\left\{ \frac{1}{\epsilon^2} \left(D_1 + \frac{D_2}{(1 - 2a)b} \right) \right\}^{\frac{1 - a}{a}}\\
	 &\quad + \frac{2(1 - a)D_2}{a^2\epsilon^2(1 - 2a)b^2}\left\{\frac{1}{\epsilon^2}\left(D_1 + \frac{D_2}{(1 - 2a)b} \right) \right\}^{\frac{1 - a}{a} - 1}\frac{D_2}{\epsilon^2(1 - 2a)b^2} \\
	 &= \frac{2(1 - a)D_2}{a^2\epsilon^2(1 - 2a)b^2}\left\{ \frac{1}{\epsilon^2}\left(D_1 + \frac{D_2}{(1 - 2a)b}\right) \right\}^{\frac{1 - a}{a} - 1}\frac{D_2}{\epsilon^2(1 - 2a)b^2} > 0,
\end{align*}
which implies that $N$ is convex. Hence, there is a critical batch size $b^{\star} = \frac{D_2(a - 1)}{aD_1(2a - 1)} > 0$.

{\bf (Decay 2):} Let $N = bK$. Then, we have that
\begin{align*}
N' 
&= K + bK' \\ 
&= \frac{(bD_1 - D_2)^2}{(b\epsilon^2 - D_2)^2} - \frac{2D_2b(bD_1+D_2)(D_1-\epsilon^2)}{(b\epsilon^2 - D_2)^3}\\ 
&= \frac{bD_1+D_2}{(b\epsilon^2 - D_2)^3}
\{(bD_1+D_2)(b \epsilon^2-D_2)-2D_2b(D_1-\epsilon^2)\}\\ 
&=\frac{bD_1+D_2}{(b\epsilon^2 - D_2)^3} \{D_1\epsilon^2 b^2+3D_2(\epsilon^2-D_1)b-D_2^2\}.
\end{align*}
If $N' = 0$, we have that $D_1 b + D_2 = 0$, i.e., $b = -\frac{D_2}{D_1} < 0$. Moreover,
\begin{align*}
	N'' &= 2K' + bK'' \\
	 &= -\frac{4D_2(bD_1+D_2)(D_1-\epsilon^2)}{(b\epsilon^2 - D_2)^3} + \frac{2D_2b(D_1-\epsilon^2)(2D_1\epsilon^2b+D_1D_2+3D_2\epsilon^2)}{(b\epsilon^2 - D_2)^4} \\
	 &= \frac{2D_2(D_1-\epsilon^2)}{(b\epsilon^2 - D_2)^4}\{-2(bD_1+D_2)(b\epsilon^2-D_2)+b(2D_1\epsilon^2b+D_1D_2+3D_2\epsilon^2)\} \\
	 &= \frac{2D_2(D_1-\epsilon^2)}{(b\epsilon^2 - D_2)^4}(3D_1D_2b+D_2\epsilon^2b+2D_2^2) > 0,
\end{align*}
which implies that $N$ is convex. We can check that $N' (b) > 0$ for all $b > \frac{D_2}{\epsilon^2}$.

{\bf (Decay 3):} Let $N = bK$. Then, we have that
\begin{align*}
	N' &= K + bK' \\
	 &= \left\{\frac{1}{\epsilon^2} \left(D_1 + \frac{2aD_2}{(2a - 1)b} \right)\right\}^{\frac{1}{1 - a}} - \frac{2aD_2}{\epsilon^2(1 - a)(2a - 1)b}\left\{\frac{1}{\epsilon^2}\left(D_1 + \frac{2aD_2}{(2a - 1)b}\right) \right\}^{\frac{1}{1 - a} - 1} \\
	 &= \left\{\frac{1}{\epsilon^2}\left(D_1 + \frac{2aD_2}{(2a - 1)b} \right)\right\}^{\frac{1}{1 - a} - 1}\left\{\frac{1}{\epsilon^2}\left(D_1 + \frac{2aD_2}{(2a - 1)b} \right) - \frac{2aD_2}{\epsilon^2(1 - a)(2a - 1)b} \right\}.
\end{align*}
If $N' = 0$, we have that
\begin{align*}
	\frac{1}{\epsilon^2}\left(D_1 + \frac{2aD_2}{(2a - 1)b} \right) - \frac{2aD_2}{\epsilon^2(1 - a)(2a - 1)b} = 0, \text{ i.e., }
	b = \frac{2a^2D_2}{(2a - 1)(1 - a)D_1}.
\end{align*}
Moreover, 
\begin{align*}
	N'' &= 2K' + bK'' \\
	 &= - \frac{2aD_2}{\epsilon^2(1 - a)(2a - 1)b^2}\left\{ \frac{1}{\epsilon^2} \left(D_1 + \frac{2aD_2}{(2a - 1)b} \right)\right\}^{\frac{a}{1 - a}}\\
	 &\quad + \frac{2aD_2}{\epsilon^2(1 - a)(2a - 1)b^2}\left\{ \frac{1}{\epsilon^2} \left(D_1 + \frac{2aD_2}{(2a - 1)b}\right) \right\}^{\frac{a}{1 - a}} \\
	 &\quad + \frac{2a^2D_2}{\epsilon^2(1 - a)^2(2a - 1)b}\left\{\frac{1}{\epsilon^2}\left(D_1 + \frac{2aD_2}{(2a - 1)b} \right)\right\}^{\frac{2a - 1}{1 - a}}\frac{2aD_2}{\epsilon^2(2a - 1)b^2} \\
	 &= \frac{2a^2D_2}{\epsilon^2(1 - a)^2(2a - 1)b}\left\{\frac{1}{\epsilon^2(D_1 + \frac{2aD_2}{(2a - 1)b})}\right\}^{\frac{2a - 1}{1 - a}}\frac{2aD_2}{\epsilon^2(2a - 1)b^2} > 0,
\end{align*}
which implies that $N$ is convex. Hence, there is a critical batch size $b^{\star} = \frac{2a^2D_2}{(2a - 1)(1 - a)D_1} > 0$. 
\qed

\subsection{Proof of Theorem \ref{theorem:4}}
\label{a_4}
Using $K$ defined in Theorem \ref{theorem:2} leads to the iteration complexity. For example, SGD using {\bf (Constant)} satisfies $N(b) = \frac{C_1 b^2}{\epsilon^2 b - C_2}$ (Theorem \ref{theorem:3}). Using the critical batch size $b^\star = \frac{2 C_2}{\epsilon^2}$ in (\ref{cbs}) leads to 
\begin{align*}
\inf
\left\{ N \colon \min_{k\in [0:K-1]} \mathbb{E}[\|\nabla f(\bm{\theta}_k)\|] \leq \epsilon \right\} \leq N(b^\star) = \frac{4 C_1 C_2}{\epsilon^4}, \text{ i.e., } 
\mathcal{N}_\epsilon = O \left(\frac{1}{\epsilon^4} \right).
\end{align*}
A similar discussion, together with using $N$ defined in Theorem \ref{theorem:3} and the critical batch size $b^\star$ in (\ref{cbs}), leads to the SFO complexities of {\bf (Decay 1)} and {\bf (Decay 3)}. Using $N$ defined in Theorem \ref{theorem:3} and a batch size $b = \frac{D_2 + 1}{\epsilon^2}$ leads to the SFO complexity of {\bf (Decay 2)}. 
\qed

\end{document}